# Architecture of Information

Yurii Parzhyn

(Augusta University, USA)

The paper explores an approach to constructing energy landscapes of a formal neuron and multilayer artificial neural networks (ANNs). Their analysis makes it possible to determine the conceptual limitations of both classification ANNs (e.g., MLP or CNN) and generative ANN models. The study of informational and thermodynamic entropy in formal neuron and ANN models leads to the conclusion about the energetic nature of informational entropy. The application of the Gibbs free energy concept allows representing the output information of ANNs as the structured part of enthalpy. Modeling ANNs as energy systems makes it possible to interpret the structure of their internal energy as an internal model of the external world, which self-organizes based on the interaction of the system's internal energy components. The control of the self-organization and evolution process of this model is carried out through an energy function (analogous to the Lyapunov function) based on reduction operators. This makes it possible to introduce a new approach to constructing self-organizing and evolutionary ANNs with direct learning, which does not require additional external algorithms. The presented research makes it possible to formulate a formal definition of information in terms of the interaction processes between the internal and external energy of the system.

## Introduction

One of the most common terms not only in science but also in the social sphere and even in everyday life is "information." As is well known, this term gained widespread recognition in scientific circles thanks to Claude Shannon's work "A Mathematical Theory of Communication" (1948). However, each science and even each individual subjectively interprets this term, often having only an intuitive understanding of it. We are accustomed to associating information with some abstract entity necessary for communication and understanding the world. It is no exaggeration to say that information, along with energy, is one of the most enigmatic concepts that touch upon many scientific disciplines: physics, chemistry, biology, philosophy, cybernetics, computer science, social disciplines, and even theology. But while the concept of energy has formal definitions in physics and chemistry, we do not encounter a formal definition of information that is universal and defines its nature. Obviously, the complexity of formalizing this concept lies precisely in its interdisciplinarity and multifacetedness. Interest in a formal definition of information is intensifying in connection with emerging problems in training large language models (LLMs) of AI and the decreasing level of trust in the information they generate [1-4].

However, what do we know about information from a formal, scientific point of view? Shannon's information theory formalizes the concept of information through entropy: information is the difference in entropy before and after receiving a message (i.e., information is a measure of entropy reduction). But his approach is limited to data processing and does not explain the nature of information, i.e., its "essence" beyond data transmission systems. Kolmogorov's algorithmic interpretation of information considers it as the complexity of an object, which can be measured using an algorithmic description—the length of the shortest program that generates it. This consideration of information does not encompass many aspects of its interaction and measurement potential, making it difficult to interpret in physical, biological, economic, and social systems.

However, the definition of information in quantum mechanics, given by John von Neumann, is already associated with the description of a system's state through entropy, which is analogous to Shannon's entropy [5]. There is also the concept by Stephen Hawking, who considers the notion of information in the context of quantum mechanics and black holes. In his works, information is understood as a characteristic of a system's quantum state and includes *parameters necessary for a complete description of the state of matter or radiation* [6].

# 1. Entropy

Thus, it can be concluded that in quantum physics, information can be interpreted as *the structure of measurable parameters of a system that describe its states or energy radiation.* For example, the Schrödinger equation describes the states of an atom using wave functions (or state functions) that depend on the coordinates of all its particles [7].

This allows for bridging the understanding of entropy in thermodynamics and quantum mechanics with the concept of Shannon's informational entropy. Let us now examine the formula for Boltzmann's thermodynamic entropy and highlight some well-known facts

$$S = k_B \ln \Omega \tag{1},$$

where $k_B$ is the Boltzmann constant, and $\Omega$ is the number of possible microstates defined by *measurable parameters*: the position in space, momentum, and energy state of each particle, under specific macroscopic conditions (states) – measurable parameters of the entire system such as energy, volume, pressure, and temperature. In other words, $\Omega$ represents the thermodynamic probability, which establishes the relationship between the microscopic states of a system and its macroscopic properties.

Let us compare this formula with Shannon's informational entropy formula (in this formula, the natural logarithm is used instead of the base-2 logarithm from Shannon's original formula to align it with the general form of the thermodynamic entropy formula)

$$H = -\sum_{i=1}^{n} p(x_i) \ln p(x_i) \tag{2},$$

where $p(x_i)$ is the probability of the occurrence of event $x_i$ in a certain random process (for example, a sequential process of receiving information from a given source).

We can see that for an ideal gas, thermodynamic probability is considered as the ratio of the number of microstates under a certain molecular distribution to the maximum number of microstates under a uniform molecular distribution, given specific macroscopic conditions (i.e., a threshold of maximum entropy is defined). This thermodynamic probability can be interpreted as the probability of an event, analogous to the frequency of a symbol appearing in a message and the probability distribution of all symbols in the message in Shannon's formula.

Thus, thermodynamic entropy $S$ can be regarded as a scaled (through Boltzmann's constant $k_B$) version of Shannon's informational entropy $H$.

In the general case, when the microstates have unequal probabilities ($p_i \neq 1/\Omega$):

$$S = -k_B \sum_i p_i \ln p_i \qquad (3).$$

If the probabilities of all microstates $p_i$ are equal ($p_i = 1/\Omega$), then Shannon entropy fully coincides with thermodynamic entropy:

$$H \propto S \qquad (4).$$

Thus, it can be assumed that Shannon entropy and Boltzmann entropy represent different perspectives on the same fundamental phenomenon.

A similar transformation can be applied to von Neumann entropy. For mixed states, this entropy is described by the following expression:

$$S(\rho) = -Tr(\rho \log \rho) \qquad (5),$$

where $\rho$ is the density matrix describing the state of the system, and $Tr$ denotes the trace of the matrix, reflecting the uncertainty in the probability distribution of quantum states:

$$S = - \sum_i p_i \, ln \, p_i \qquad (6),$$

where $p_i$ are the eigenvalues of the density matrix $\rho$.

Thus, thermodynamic entropy and quantum entropy could be considered as special cases of the more general concept of Shannon's informational entropy, with additional physical interpretations. The relationship between thermodynamic and informational entropy remains a subject of modern research and scientific discussions [8].

However, the influence of information on entropy reduction is just *one of its properties, not its essence*. This property clearly indicates the energetic nature of information. Our understanding of this connection is further reinforced by recalling Landauer's principle [9]: the erasure of information in a computational system is associated with a physical energy cost, leading to heat dissipation into the surrounding environment:

$$E = k_B T \ln 2 \qquad (7),$$

where $k_B$ is the Boltzmann constant, $T$ is the environmental temperature (in kelvins), and ln2 is the logarithm of the number of states involved in erasing one bit of information.

Thus, according to this intuitively understandable principle, the erasure of information increases the entropy of the external environment, compensating for the reduction of entropy within the system. And although Landauer's principle is largely conceptual and its physical manifestation is subject to various limitations, it can still be perceived as indirect evidence of the fundamental connection between energy and information.

But let us return to Shannon's informational entropy. In this formula, entropy is considered as the average amount of information, measured in bits, that is necessary to describe (or represent using a binary number system) a random event or a sequence of random events. The less probable an event is, the higher its entropy, and the more information (quantified in bits or other structural units of an informational sequence) it carries—or rather, the more information is required to resolve its uncertainty. Consequently, such an event has a higher informational value and requires a greater number of conditional bits for its representation in the system.

We can now take an important step in our reasoning. If, following Landauer, we consider the value of an informational bit as a measurable value of energy (or a measurable parameter of energy), then the probability of this value in Shannon's formula takes on an entirely different meaning: this probability reflects the relationship or *interdependence of measurable energy parameters within the spatiotemporal structure of signals received (or perceived) by the system*. Furthermore, considering that in quantum physics and chemistry, the relationships between structural elements of energy also constitute energy—namely, the energy of interaction—then the probability of an event in Shannon's formula can be *interpreted as the degree of our (or the system's) ignorance regarding the values of these interaction energy parameters*.

The transition from the probabilistic relationship of event sequences in the definition of informational entropy to the energetic relationship of sequences of measurable parameters of energy elements can be demonstrated as follows:

$$(x_1 p_{12} x_2 \cdots x_{n-1} p_{n-1,n} x_n) \leftrightarrow (e_1 \gamma_{12} e_2 \cdots e_{n-1} \gamma_{n-1,n} e_n) \qquad (8),$$

where $x_i$ is an event or a bit of information in the binary number system, $p_{ij}$ is the conditional probability of two consecutive events in the structure of the received message, and $e_i$ is the measurable parameter of an energy element (which, in Shannon's formula, can take the values 0 or 1). The term $\gamma_{ij}$ represents the probability of the relationship between two consecutive energy elements, reflecting the level of external (relative to the energy-emitting system) understanding of this relationship or the degree of confidence in its existence.

We can conditionally perceive this probability $\gamma_{ij}$ as an abstract equivalent of the interaction energy of structural energy elements. This approach is convenient because introducing such an equivalent allows it to be interpreted as a representation of the internal energy of the system under consideration, which, in turn, enables the construction of models describing the changes in this energy.

A similar analogy can be found in generative artificial neural networks, where $\gamma_{ij}$ can be interpreted as the weight coefficients representing the relationship between tokens. However, it is important to note that this probabilistic abstraction only reflects the surface-level connection between measurable energy parameters—more precisely, their statistical relationship.

The validity of using the probability $\gamma_{ij}$ in this formula to describe the interaction energy is supported by the analogous probabilistic representation of this energy in the Hartree-Fock method in quantum chemistry, as well as in the description of the wave function, quantum operators and their eigenvalues, the tunneling effect, and Heisenberg's uncertainty principle in quantum physics [10, 11]. Furthermore, this approach is also employed in molecular dynamics and statistical mechanics.

This interpretation of probability in Shannon's formula could also serve as a foundation for bridging his informational entropy with algorithmic entropy—Kolmogorov complexity [12].

Let us highlight another important point. In the proposed interpretation of Shannon's formula, $x_i$ should not be understood as a measurable parameter of an energy element but rather as *a generalized representation of an energy value within a certain symbolic system, i.e., an abstraction*.

Considering that the same or different types and values of energy can be represented in various symbolic systems—for example, in the binary system, they correspond to bit values, while in the decimal system, they may correspond to probability values—this representation is *inherently subjective*.

Thus, in the abstract formulation of the information entropy equation, $x_i$, $p_{ij}$ or $e_i$, $\gamma_{ij}$ acquire the meaning of elements of the alphabet of an information system. These elements can be assigned different meanings by either us or the system, but they lack an internal structure that could be directly associated with the structure of energy.

Let us summarize our considerations regarding Shannon's information entropy formula:

1. The informational elements in the formula are represented as abstract elements of the alphabet of an information transmission and processing system (e.g., in bits or words) rather than as measurable parameters of the system's energy (micro- or macrostates) that formed it. This abstraction does not allow for an examination of *the internal structure or essence of these informational elements*, which is the reason for the absence of their contextual (deep) interconnection. For example, measurable energy parameters are interdependent, and this interdependence is determined by the structure of the system's internal energy. This is clearly observed in quantum physics and chemistry.

2. The informational elements form *a linear spatiotemporal structure*. This reflects the process of information processing and transmission in informational systems but significantly simplifies *the structure of real relationships* between informational or energetic elements.

3. The interconnection of informational elements in the formula is considered only at *an external, statistical level*, which does not reflect *the deep interdependence of measurable energy (information) parameters*—the dependence of macrostates on the microstates of the system.

These limitations lead to the emergence of a phenomenon known as the "*entropy gap*." This term can be discussed in relation to the emerging challenges of generative artificial neural networks (ANNs) [13]. The essence of this phenomenon does not lie in the "jump" of reduced information entropy at the output of an ANN (which is a natural process) but in the fact that the decrease in information entropy is associated with an increase in the probability of interconnection between units of information in the learning process, meaning it is based on the statistical nature of informational relationships. This approach forms the foundation of the modern connectionist concept of ANN construction and training.

However, the most informative relationships are precisely those with low probability, i.e., statistically rare connections. This prevents modern ANNs from generating new or "creative" information. Everything that an ANN creates (generates) is statistically the most probable, meaning it depends on the volume of data used for training. However, from the perspective of novelty and creativity, this results in averaged data. Thus, rare relationships, which could drive innovative solutions or new ideas, do not receive adequate weight because their contribution to the overall probability distribution is minimal. As a result, the model's output becomes predictable and constrained by averaged statistical patterns. This is where the "entropy gap" manifests—between creativity and mediocrity.

All of this points to the conceptual limitations of the connectionist approach in constructing and training ANNs, as any attempts to account for reduced statistical interconnection within the entropy framework will inevitably lead to an increase in "chaos." This means that the level of trust in such generated information will decline.

Some AI researchers argue that the available data for training AI models have been exhausted and link the further development of models to the use of synthetic data generated by the models themselves. However, information entropy clearly indicates that such an approach, based on data inbreeding, will quickly lead to information degradation and model collapse, exhibiting a negative effect similar to overfitting [14]. Moreover, attempts to overcome conceptual limitations within the same framework may bring some improvements but will ultimately lead to a global failure. *To effectively address the emerging challenges, a paradigm shift is necessary—one that redefines information and its role in the process of system construction.*

Based on these considerations, we could formulate a definition of information that unifies the concepts of thermodynamic and informational entropy—as *a generalized subjective representation of energy within a given symbolic system*. However, such a definition is largely philosophical and lacks sufficient formalization for a rigorous scientific concept.

## 2. Energy Landscape

The idea of converging the understanding of the general nature of energy and information can be realized by applying the concept of *the energy landscape* to both energetic (thermodynamic, dissipative, quantum, etc.) and informational, biological, economic, or social systems.

In physics, an energy landscape is a concept that describes how a system evolves in state space, striving toward a state of minimal energy. For example, in mechanics, an energy landscape can be a representation of a system's potential energy as a function of its spatial coordinates, while in thermodynamics, it reflects free energy as a function of system parameters (such as temperature, pressure, concentration, etc.). In such landscapes, the minima—either local or global—correspond to stable states of the system, while maxima and saddle points represent unstable states. The differences between global and local minima determine the complexity of the system's dynamics or evolution.

The concept of an energy landscape is widely used in chemistry and biophysics (e.g., in protein folding and catalysis), as well as in statistical and quantum mechanics. Undoubtedly, real physical systems possess a complex energy landscape that describes the system's energy across various states.

In the general case, the total energy of a system, $U_{total}$, includes both the local energies of $i$-th structural elements (units) of the system $U_{local,i}$, and the interaction energies between units $i$ and $j$, denoted as $U_{bond,ij}$:

$$U_{total} = \sum_i U_{local,i} + \sum_{i<j} U_{bond,ij} \qquad (9).$$

The energy landscape describes the dependence of the total energy $U_{total}$ on the configuration of the system. The bonding energy $U_{bond}$ characterizes the interactions between system elements and plays a key role in shaping the energy landscape. Examples of such interactions include: covalent or ionic bonds in chemical compounds, Van der Waals intermolecular forces, elastic interaction energy in solids (e.g., in crystals or molecular lattices), and the energy of magnetic or electric dipoles in systems with magnetic moments or polar molecules. These interactions are often nonlinear, and their energy depends on the distance between structural points.

Figure 1 presents the general structure of the system's energy landscape, which is described by equation (9).

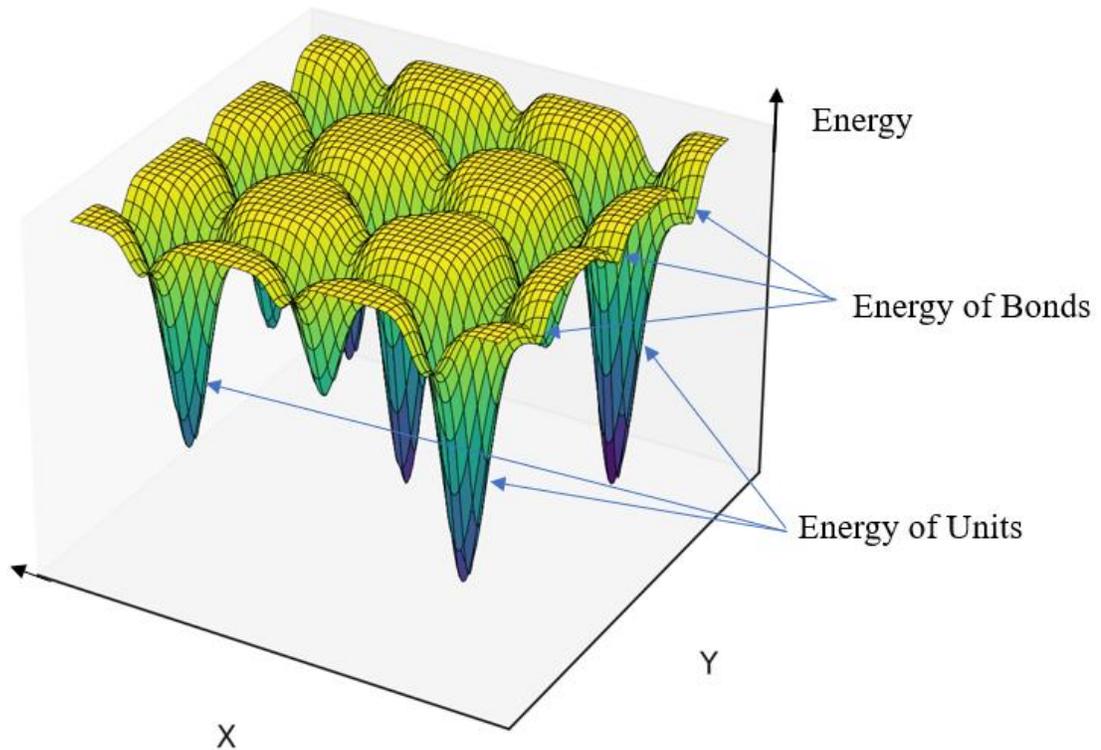

Figure 1. General structure of the energy landscape of the system.

In the general case, *a stable internal energy landscape El* of an open energy system is defined as the distribution of internal energy across structural elements (units) $U_i$ and the connections between them $W_{ij}$, without considering the processes of irreversible heat dissipation $\Delta T$, which are associated with energy fluctuations of the units around their stable state, and without accounting for external energy or energy fluxes between the system and the environment.

$$El_{rest} = min(\sum_{i=1}^{N} U_i + \sum_{i=1}^{N} \sum_{j=1}^{N} W_{ij}) = Tr1 \qquad (10).$$

In this formulation, we refer to this energy landscape as *the landscape of the system at rest or the energy equilibrium landscape $El_{rest}$*, and the total energy of the system in this state as *the rest threshold Tr1*. In this state, the system reaches *maximum entropy* for a given macrostate, where no energy, mass, or substance flows are observed. For example, in thermodynamic equilibrium, all thermodynamic forces, such as temperature, pressure, or chemical potential gradients, are zero, and all internal processes that could occur within the system (e.g., chemical reactions, heat transfer) are complete.

Even if energy fluctuations at the microscopic level continue—such as changes in internal energy parameters (e.g., particle motion)—the system remains in an *equilibrium (rest) state*. This state defines the maximum number of possible microstates for a given macrostate that do not lead to its change. In other words, when the structure stabilizes and its internal energy is minimized, we describe the system as having reached its maximum entropy level for a given macrostate.

The rest or equilibrium state of a system is an idealized concept for a closed thermodynamic system. When such a system reaches a state of rest, it possesses maximum entropy.

If we consider an open system that continuously receives external energy, allowing it to maintain a non-equilibrium state and create ordered structures, such a system—referred to as a dissipative system—experiences *a local decrease in its internal entropy due to structuring, leading to the formation of dissipative structures*. At the same time, the overall macrostate of the system may remain unchanged, which still results in an increase in total thermodynamic entropy, in accordance with the second law of thermodynamics. These processes form the foundation of Prigogine's theory of dissipative systems [15].

Examples of systems in a rest state, depending on the interpretation of energy, include: a gas in a closed, isolated vessel that has reached thermodynamic equilibrium (where temperature, pressure, and density are uniform throughout the volume); a molecule in vacuum; an isolated biological neuron in a resting state; an artificial neural network in a resting state (i.e., in the absence of input signals and changes in internal parameters).

Analogies (metaphors) can also be drawn for economic or social energy at rest, such as a company's financial reserves stored in a bank or a society's system of cultural and moral values.

However, it is important to note that this state indicates a fundamental limit—*the rest energy threshold of the system, which defines the minimum internal energy of the system and the maximum entropy* for a given structure and a specific set of macrostates. The energy landscape of a system at rest depends on its internal physical structure. Thus, this landscape forms the energetic foundation of the system, which determines its resilience to external energetic influences.

The minimum energy of the rest-state landscape is ensured by *minimizing the physical structure of the system and/or minimizing the internal (potential) energy of the units and their interconnections*. Structural minimization can be associated with its dynamic adaptation in response to changes in external energy influences or with evolutionary structural changes, as observed in bifurcation points of dissipative systems or in biological systems.

However, if we speak about minimizing the internal energy of the units and their interconnections, we must recognize that *this energy itself has a structure*. Examples of describing such energy structures can be found in quantum mechanics and quantum chemistry, such as Schrödinger's equation, the Hartree-Fock method, the Lennard-Jones potential, and density functional theory (DFT), among others.

From quantum mechanics, the bond energy $E_{bond}$ reduces the total energy of individual atoms $E_{atoms}$ within the overall molecular energy structure $E_{molecule}$:

$$E_{bond} = \sum E_{atoms} - \sum E_{molecule} \qquad (11).$$

This implies that part of the potential energy of the units is used to create interconnections between them. Thus, the energy landscape of the system at rest consists of "wells" of potential energy, represented by the units, and "channels" for potential energy transitions between wells, represented by interconnections (Figure 1). The potential energy within these wells and channels has an internal structure, consisting of energy levels or sublevels.

These potentials effectively determine the energy capacity of the system and indicate the "bottom level" of this capacity relative to some reference point (the "plateau" of the landscape). For example, in an atom, the potential energy of an electron in the nucleus's field defines the "bottom" of its energy state relative to the nucleus's energy. Metaphorically, we can visualize this level as the geodetic height of a point in an energy landscape above sea level.

*Thus, the formation of stable structures leads to a decrease in the internal energy of the system in a resting state.*

If we apply this description of the rest-state energy landscape to ANNs, we see that:

1. For most types of models, the rest-state energy landscape is static and, as it might seem, consists solely of weight coefficients, which we can interpret as the energies of interconnections between neurons (units). In these models, the energy is considered in only one direction for signal propagation. However, this is not the case, because in ANNs, the interconnections exist "physically," but they do not exist as distinct energy elements. If we look at the McCulloch-Pitts model, which forms the foundation for most neural models in ANNs:

$$Y = f(\sum_{i=1}^{n} w_i x_i + Q) \tag{12},$$

where: $Y$ is the neuron's output (response), $f$ is the activation function, $w_i$ is the weight coefficient of the $i$-th neuron input, $x_i$ is the input signal to the $i$-th input, $Q$ is the activation threshold of the neuron. When, we see that the weight coefficients serve as modulators of the input signals (input energy) of the neuron, rather than independent energy elements of interconnections. However, if we consider that the input signals correspond to the values of bits (as in the classical model), then from an energy perspective, the sum of the weight coefficients could determine the energy structure and the possible depth of the potential "well" of the unit's energy landscape in the ANN and, consequently, the overall energy landscape of the system in its resting state (10). That is

$$U_{jrest} = \sum_{i=1}^{N} w_{ij} \tag{13},$$

$$El_{rest} = \sum_{j=1}^{N} U_{jrest} = \sum_{i=1}^{N} \sum_{j=1}^{N} W_{ij} \tag{14},$$

where $U_{jrest}$ is the energy of the $j$-th unit (neuron) in the resting state.

Indeed, if the input $x_i = 0$, then this input energy does not leave the $i$-th sublevel of the unit's potential "well," with $w_{ij}x_i = 0$. However, if $x_i = 1$, the energy output from this sublevel equals $w_{ij}$, meaning that the sum of weight coefficients in this case determines the maximum "depth" of the potential "well."

In general, drawing an analogy with an energy system, such as an atom, we could compare the operation of a neuron to the ionization process of an atom, where the linear combination of inputs $\sum w_i x_i + Q$ models the release of an electron from the atom, i.e., the generation of free energy. Meanwhile, $w_i x_i$ models the transition of an electron to another energy sublevel, where $w_i$ defines the energy of the sublevel in the resting state. However, this is merely a conditional analogy that, in this case, lacks a common energetic foundation.

Thus, the formal neuron model is an informational model based on the mathematical, or more precisely, statistical abstraction of information processes and, in its classical interpretation, is far from describing thermodynamic processes.

2. The values of individual weight coefficients in most ANN models could be interpreted as a conditional internal energy of the system, which lacks structure and is not connected to the system's input energy. Indeed, each neuron simply distributes (duplicates) its output energy evenly across all neurons in the next layer. This energy distribution can be compared to its dissipation. However, in reality, such "dissipation" in an ANN (if one considers its physical implementation) is associated with additional energy costs required to maintain the energy level of the neuron's response at the inputs of multiple neurons in the next layer.

The weight coefficients of inputs are associated with a different type of energy—pseudo-energy. However, this pseudo-energy does not govern the distribution of external energy across the system's internal energy landscape, as it might seem. Instead, it determines the coordinated potential contributions of individual elements of the neurons' input energy, directed at overcoming their activation thresholds. This pseudo-energy represents the error energy, which is minimized during the iterative learning process.

Thus, the weight coefficient formally defines the "correct" contribution of input energy to overcoming the activation threshold, from the perspective of minimizing error energy. In other words, it determines the release of energy from the "potential well" to achieve the required response of both the neuron and the ANN as a whole.

3. The input energy of an ANN is normalized and constrained by a representation system, which can conditionally be compared to sensory perception that segments and transforms external energy into internal energy. Thus, both input and output energies are represented as another type of pseudo-energy, *expressed in the symbols of the system's internal alphabet*, for example, in binary or decimal numerals.

Such a representation of input energy, along with other specified constraints, leads to the formation of a deterministic set of possible *microstates of the system*. In ANNs, these microstates are determined by the sets of weight coefficients in the resting state and the distribution of input energy across the network, i.e., the responses of neurons during information processing.

The set of *macrostates* of an ANN is then defined by the responses of the output layer neurons and, from an energetic perspective, forms the system's free energy. In this sense, the network's output reflects a new stable state of the system (in which it has minimized its energy), conditioned by the interaction of input energy (the input signal vector) with the energy landscape.

One could say that an ANN in the inference state (the operation of a trained network) is in a state of *stable energetic nonequilibrium*. In this state, the system exits its resting state, but the structure of its internal energy landscape remains unchanged. The system maintains a stable internal (physical) structure and dynamics, as well as predefined macrostates.

This state is associated with the existence of a second fundamental threshold—the system's activation threshold $Tr2$. This threshold defines the system's internal metric, allowing for the determination of the depth of "wells" and "channels" in the system's energy landscape. The magnitude of a unit's local activation threshold determines the *depth of its potential energy "well," i.e., its energy capacity, and depends on the internal structure of its energy.*

For example, at the quantum level, an electron must overcome several energy sublevels to detach from an atom, meaning it must surpass the ionization barrier, which can be considered an analogue of the activation threshold. Different systems will have different activation thresholds: for biological neurons, it is the action potential; for an artificial neuron, it is the activation function. In dynamic systems, the activation threshold may be described by Lyapunov function or functional. Depending on the type of system, the complexity of interactions, and system properties, other methods may be used, such as Jacobian matrices, phase diagrams, energy barriers, feedback mechanisms, statistical methods, critical phenomena, and more.

Thus, the type and structure of internal energy define the rules or conditions for the system's activation. In the general case:

$$E_{act(i)} = \left(U_{min(i)} + \Delta U_i\right) > Tr2_i \qquad (15),$$

$$\Delta U_i = F(X) \qquad (16),$$

where: $E_{act(i)}$ – activation energy of the *i*-th unit, $U_{min(i)}$ – potential resting energy of the *i*-th unit, $\Delta U_i$ – additional energy of the *i*-th unit, $Tr2_i$ – activation threshold of the *i*-th unit – the limit of its potential capacity, $F(X)$ – function of input energy distribution across units.

Let us imagine a surface with pits connected by hollows or channels, and possibly hills. If we pour water onto this surface, we will observe the process of energy distribution across the energy landscape of the system, filling the pits and channels. This representation vividly illustrates the process of energy distribution across the energy landscape of the system.

If there is an influx of energy into the system and no effective outflow, the energy will overflow the system's capacity, and the energy connections between the units may be destroyed, as their potential capacity is much smaller than that of the units themselves. This connection capacity determines the third fundamental threshold of structural stability or system resilience $Tr3$.

$$E_{str(ij)} = \left(W_{min(ij)} + \Delta U\right) \leq Tr3_{ij} \qquad (17),$$

where: $E_{str(ij)}$ – energy of the stable connection between units $i$ and $j$, $W_{min(ij)}$ – minimum potential energy of the connection between units $i$ and $j$, $Tr3_{ij}$ – threshold of structural stability for this connection.

If the destruction of connections exceeds the structural stability threshold of the entire system, this may define the bifurcation point of the system, i.e., the point of its state change or further development, as well illustrated in Prigogine's theory of dissipative systems.

In an ANN, the activation threshold is represented by a rule or function that defines the artificial neuron's response. In most modern neuron models, this activation function (e.g., sigmoid or ReLU) is related to response normalization and does not reflect the level of threshold exceedance. However, it can serve as an analog of the neuron's potential energy level in the activated state.

The existence of two thresholds, $Tr1$ and $Tr2$, determines the stability of any system against fluctuations within the inter-threshold space. Exceeding the activation threshold by a certain magnitude results in the same stable system response, for example, at the system's output. To transition the system into an unstable state, external energy is required to alter the resting energy, the activation threshold, or to surpass the structural stability threshold.

Since the energy landscape consists of multiple components, an energy fluctuation that disrupts local elements of the $Tr3$ threshold may not lead to the complete collapse of the entire landscape. However, an increase in the number of such fluctuations can bring the system to a bifurcation point.

This approach to analyzing state changes in the system's energy landscape offers a new perspective on system classification:

- *If the system can adapt to changing conditions by modifying its thresholds or internal energy parameters, it is classified as adaptive.*

- *If adaptation occurs through the restructuring of its resting energy landscape, which involves changing its internal structure, the system is evolutionary.*

- *If the system can create a new resting energy landscape by synthesizing its internal structure, it is classified as self-organizing.*

Thus, the activation threshold defines the maximum allowable fluctuation of internal energy for maintaining a resting (equilibrium) state. If this limit—determined by the structure of internal energy—is exceeded, the system transitions into a stable nonequilibrium state.

The bifurcation threshold depends on the system's structure, its resting and activation thresholds. If system response stability deteriorates with increasing fluctuations, this may indicate proximity to the bifurcation threshold.

The system may then transition either to adaptation through self-organization or to a state of degradation, meaning a nonequilibrium state with a disruption of its internal energetic and physical structure.

An important conclusion can be drawn: *the structure of energy serves as the foundation not only for shaping the system's internal energy landscape but also for its processes of adaptation, evolution, and self-organization.*

If we consider that system adaptation can occur not only through feedback but also through deep internal mechanisms of energy transformation, which involve changes in thresholds and the energy landscape, this could significantly expand system theory and cybernetics. Moreover, it could transform our understanding of learning in artificial neural networks, paving the way for a new perspective on their self-learning capabilities.

## 3. Hopfield Model

The first and most well-known ANN model that utilizes the concept of an energy landscape and an energy function is the Hopfield model [16, 17]. This model represents an attempt to bridge the gap between information theory and physics. Let us analyze how well-founded this convergence is.

From the perspective of the energy states in the model, the most well-known physical interpretations of Hopfield's concept assert the following:

1. The energy landscape is interpreted as the set of all possible states of neurons.

2. The minimization (or rather, reduction) of energy during state changes, both in individual neurons and in the network as a whole, is illustrated by the Hopfield energy function.

3. A local attractor, which defines the "correct" state of a single neuron as an element of a global attractor, has a "basin of attraction". This basin is determined by the states of the neuron that converge toward a stable value, which corresponds to a stored pattern element. Failure of a neuron to reach the local attractor state is characterized as an undesirable phenomenon—a "trap" in a local energy minimum. This effect can arise due to interference between stored patterns.

4. Based on these assertions, it follows that there exists a set of global attractors with minimal energy, each of which corresponds to a stored pattern.

As a prototype for the Hopfield model, the Ising model is used to describe the properties of ferromagnetic materials [18]. The Ising model is a physical model, and its primary characteristics include energy, entropy, and temperature, which are used to describe the behavior of a system consisting of a large number of interacting spins in magnetic materials. In these materials, interactions between magnetic moments (spins) lead to complex and disordered states.

Spins represent the orientation of atomic magnetic moments and are associated with the energy of their interactions. The Ising model was developed to understand how local interactions between neighboring spins can lead to global ordering in systems, such as the phase transition of a ferromagnet. This ordering depends on temperature: at low temperatures, spins tend to align in the same direction (ordered state), and at high temperatures, spins become randomly oriented (disordered state). This model does not account for all forms of energy, nor does it describe energy dissipation or free energy formation, meaning it is not a complete thermodynamic model.

If we analyze the model in more detail, it studies the *collective behavior of spins $S_i$, which can take values* ±1. The interaction between spins is described by a matrix of constant or random values $J_{ij}$. This $J_{ij}$ can be interpreted as t*he conditional interaction energy, which can reflect spin correlation: a positive value corresponds to parallel spin orientation, or a negative value corresponds to spin misalignment (disorientation)*.

It is crucial to note that $J_{ij}$ represents the interaction energy rather than the internal energy of a single unit (atom). However, in the classical Ising model, all $J_{ij}$ values are set to +1, meaning it is axiomatically assumed that spin states tend to converge toward a single stable state with parallel alignment.

The current states of spins depend on temperature, which influences the probability of spin reorientation according to the Boltzmann distribution: the lower the temperature, the higher the probability of parallel spin alignment.

During the dynamic change of spin states, a complex energy landscape of the system is formed. The dynamics of the Ising model are described through the energy functional, the Ising Hamiltonian $H$:

$$H = -\frac{1}{2}\sum_{i \neq j} J_{ij}S_iS_j - \sum_i h_iS_i \tag{18},$$

where $h_i$ is the external field.

As the system evolves, its energy decreases until it reaches a minimum, characterizing a stable state with parallel spin orientations. In the absence of an external field ($h_i = 0$), this state is fully defined by the matrix of spin interaction weights. Thus, i*n the stable equilibrium state of the system, the energy landscape, according to equation (18), is represented by the weight matrix with values of +1,* reflecting the state of the system with minimum energy and maximum thermodynamic entropy.

The Ising Hamiltonian is a discrete analog of the Lyapunov function, which is used to describe continuous stable states of dynamic systems [19].

In the Hopfield model, the Ising model's concept is adapted to solve the problem of building associative memory. The energy function of the Hopfield model, which decreases with each step of neuron state updates, is described by the following functional:

$$E \; = \; -\frac{1}{2}\sum_{i \neq j} w_{ij}s_i s_j - \sum_i \theta_i s_i \qquad (19),$$

where $s_i$ are the neuron states (+1 or -1), $w_{ij}$ are the connection weights between neurons, and $\theta_i$ are the threshold values.

One of the key differences between these models is that in the classical Ising model, the changes in spin states occur randomly using the Metropolis algorithm (a Monte Carlo method) [20], which models the relationship between the external and internal energy of the system. In contrast, in the Hopfield model, changes in neuron states occur deterministically based on comparing input signals with the signs of mutual correlations, represented by the sum of connection weights. The changes in neuron states aim to restore a specific state (pattern) due to the recurrent architecture of the model. That is, the architecture of recurrent connections in the Hopfield network acts as a sort of "broom" to maintain its dynamics.

In the Ising model, the threshold for the spin "flip" depends on the temperature $T$ and the change in energy $\Delta E$. This threshold is defined by the probability $P$ with which the system "decides" whether to adopt a new state that might have a higher energy:

$$P(\Delta E) \propto e^{-\Delta E / k_B T} \tag{20},$$

where $k_B$ is the Boltzmann constant.

The probability $P$, according to the Boltzmann distribution, serves as an analog to the energy threshold that determines the change in the spin state. *Therefore, the relationship between the energy threshold and thermodynamic parameters defines the physical basis of the Ising model.* We can see that in the Ising model, this threshold is dynamic, while in the classical Hopfield model, the threshold $\theta_i$ for changing the state of the neuron is static and equal to 0. *This threshold, along with other parameters of the Hopfield model, is not related to thermodynamics, which defines the Hopfield model as informational.*

Let us pay attention to the fact that the Hopfield model uses the concept of the system's "energy landscape" as the set of the neurons' states, but does not provide its definition for the equilibrium resting state. However, if this is an energetic system, then there must exist a landscape in the resting state.

It is easy to see that, according to the Hopfield energy function, the energy landscape of the system in the resting state is formed by the sum of the weight coefficients at the neuron inputs. In other words, the weight coefficients acquire the meaning of *the internal energy of the network in the resting or equilibrium state*. Indeed, according to the Hebbian learning rule, these weight coefficients are formed once during the learning process and are not dynamically updated afterward:

$$w_{ij} = \frac{1}{N} \sum_{k=1}^{p} x_i^k x_j^k \tag{21},$$

where $x_i^k$ is the response value of the $i$-th neuron in the $k$-th sample, $N$ is the number of neurons in the network, and $p$ is the number of patterns to be memorized. Thus, in the Hopfield model, the weights $w_{ij}$ are formed based on the correlation of patterns that the network must memorize, rather than on the basis of fundamental energy interactions.

Then, according to formula (19), the minimum energy of the network in the equilibrium state $E_{min}$, i.e., theoretically, if all the states of the neurons (their responses) were perfectly correlated with each other based on their weight coefficients, will be:

$$E_{min} = -\frac{1}{2} \sum_{i \neq j} w_{ij} \qquad (22).$$

Thus, the network has *one global minimum energy in the resting state*, which is a theoretical limit, and practically, in a working network, this is not achievable due to the various types of correlations for different signals and the peculiarities of the weight coefficient formation algorithm. One can conclude that, indeed, for each memorized pattern, there will be its own minimum value of the energy function, which, in practice, will depend on the degree of distortion of the memorized patterns. This can be interpreted as *the existence of global energy minima for each memorized pattern*.

This is explained by the fact that the signs of the sums of the weight coefficients of a neuron indicate the "correct" or "desired" correlation of the input signals with the memorized patterns—either positive or negative. The sum of the weighted input signals of a neuron can be interpreted as a "*correlational vote*," as a result of which a decision is made to change or maintain the current state of the neuron. The result of this "voting," reflecting the maximum possible correlation of signals, will depend not only on the values of the weight coefficients but also on the current states of the neurons.

If complete correlation according to the weight coefficients is not achieved, then the system will not reach the desired global minimum corresponding to the memorized pattern, which will be reflected in the energy formula as an undesirable phenomenon of a local energy minimum. Therefore, in the Hopfield model, *the energy function can increase at the level of individual neurons if local correlations are disrupted*. However, the global dynamics of the network, taking into account possible constraints, will aim to reduce the overall energy function. This emphasizes that the Hopfield model works with a global energy function, and local deviations in individual elements can be compensated by the overall behavior of the system.

Thus, the following conclusions can be drawn:

1. The Hopfield model, unlike the Ising model, is informational, meaning it is abstracted from any thermodynamic or other energetic interpretation. Therefore, when considering it, it would be more appropriate to use the term "conditional" or "pseudo" energy. However, the use of concepts such as "energy landscape" and "energy function" still represents powerful conceptual steps. *The minimum of the Hopfield energy function demonstrates the attainment of a stable state and a reduction in the system's informational entropy*.

2. The classical Hopfield model has a spatially linear energy landscape of the system in its resting state, represented by the sum of the weight coefficients of the neurons. This is explained by the physical structure of the network. In the Ising model, this landscape reflects the lattice structure of atoms. When considering the Hopfield model, a dilemma arises regarding the interpretation of the energy of the weights: they could be attributed to the elements of the internal energy of neurons (Equation (10)) or to the energy of neural interactions (Equation (14)), since they are a function of the reference values of their internal states according to Equation (21). However, due to the structure of the McCulloch-Pitts model (Equations (12), (13), (14)), they should still be attributed to the elements of the internal energy of the units (neurons). In this case, the coefficient ½ in the Hopfield energy formula (Equation (19)) is redundant, unlike in the Ising Hamiltonian (Equation (18)), where the weight coefficients indeed represent the interaction energy of atoms.

3. The Hopfield energy function is separate from the process of governing the network's dynamics and serves only as an illustration of the process. Therefore, the metaphor of "attractors," whether informational or energetic, is incorrect, as there is no energy or energetic state in the model that "attracts" other energetic or informational states. That is, the figurative representation of the network's operation as a ball moving along the surface of an energy landscape and falling into a "well" with minimal energy does not correspond to real processes.

This assertion is justified by the fact that the metaphor of an "element of a global attractor" could only apply to the sum of the weight coefficients of each neuron as the only explicit, rather than virtual, constant value. However, the problem is that this sum serves the same role for all "global attractors," and all correlations between "elements of the global attractor" are interconnected by the matrix of these weight coefficients. The decision to change the state of a neuron depends on the values of incoming signals, meaning the "attractor" must be dynamic and change depending on the input signals. It should be noted that in dynamical systems theory and nonlinear dynamics, there are concepts that describe such phenomena, but for this, a system must have: multistable states and a switching mechanism between them; or attractor metamorphosis with gradual or sudden changes in the shape or type of the attractor; or chaotic evolution of the attractor, among others, which is possible in complex systems with changing parameters or interactions.

Therefore, it is more appropriate to consider not the metaphor of an "attractor" but the true correlational nature of these weight coefficients and their sum. This process can be called *correlational voting*, where the majority of weighted input signals "vote" for a positive or negative correlation of the output signal, i.e., for making a decision to transition to state +1 or -1.

4. If the energy function of the Hopfield model (19) indicates a decrease in conditional energy associated with a reduction in informational entropy, then the analogous energy function of the Ising model (18) also points to this. Thus, it can be concluded that *a physical spin-glass system produces not only thermodynamic but also informational entropy*. This is consistent with the theory of quantum informational entropy and Prigogine's theory of dissipative systems.

Overall, the functioning of the Hopfield model could be described more simply and accurately without using energetic terminology: the final stable state of the system depends on the initial state (input signal vector) and the correlation matrix defined by the weight coefficients of correlation. The process of state changes in the system is iterative and determined by recurrent connections. The final state of the system is characterized by achieving the maximum possible (stable) correlation of neuron responses (their states) in accordance with their weight coefficients. The energy function (Lyapunov function) in this interpretation takes on a demonstrative meaning and reflects the system's tendency toward a stable state or a reduction in internal informational entropy.

However, if we decide to consider the operation of this network purely from an energetic perspective, let us note that the internal energy of the network, as we have established, corresponds to the energy of neurons (units) according to formulas (13, 14), whereas the energy of neuron interconnections remains undefined. Suppose the system "does not know" what a bit of information is and instead treats it as a value of potential energy that needs to be "disposed of" in some way: stored, distributed, dissipated, or converted into output energy. Then, the dynamics of neuron state changes are determined solely by elements of external energy entering the system. In our case, this energy corresponds to the symbols $x_i$ of the bipolar alphabet (+1 – maximum energy, or -1 – minimum energy). Indeed, if no signals are supplied to the network's input, then the output of each neuron will either be zero or contain random energy values.

Thus, the values of the internal energy of the network, expressed in the form of the weight coefficient matrix $w_{ij}$, define the threshold values of the system's resting state. The dynamics of state changes in the system are determined by the excess of external energy over these resting threshold values within the threshold activation value $\theta_i$. This process should be viewed not as the reduction of the system's internal energy to a minimal (stable) level but rather as its redistribution among local points of the energy landscape (units).

An important feature of the model is that the energy landscape, determined by the weight coefficient matrix, is formed based on samples of external energy that need to be memorized, rather than in the form of conditional error energy, as in a multilayer perceptron, for example. Thus, the energy landscape of the Hopfield model could be considered as an *internal model of the external world perceived by the system*.

Thus, the process of iterative redistribution of external energy across the local levels of the energy landscape is represented in the form of a Lyapunov function, expressed through the Hopfield energy function (19). The result of this redistribution indicates the level of external energy exceeding the internal resting-state energy at the local extrema of the landscape. After comparing this excess with the threshold value $\theta_i$ of the functional, it is reflected in the system's output signals. Therefore, $\theta_i$ serves as the decision threshold for redistributing external energy among the network elements.

According to Hopfield's formula (19), with each step of energy redistribution, the total conditional energy $E$ of the network decreases, which, in accordance with the laws of thermodynamics, reflects an increase in the system's thermodynamic entropy. That is, external energy diminishes as it is "dissipated" among neurons and fills the "wells" of potential energy in the units (neurons) until a state is reached where further changes become impossible, and the system stabilizes.

Thus, we could also conclude that the Hopfield energy formula represents the dynamics of entropy change, inherently linking informational and thermodynamic entropy. However, a drawback of this description is that neurons must "dissipate" the accumulated potential energy in their "wells" before the next state update cycle. As a result, to maintain dynamic operation, the system will continuously require additional external energy.

One of the conceptual problems of both the classical Hopfield model and this alternative energetic description of its operation is the absence of a self-halting mechanism. The system itself will continue running indefinitely without state changes (in an update loop) until an external stop signal is received.

Thus, the Hopfield network, as both an energetic and informational system, can be classified as a system with *a deterministic static structure, a deterministic static generalized energy landscape in the resting state (formed instantly rather than iteratively), and a deterministic dynamic energy landscape during the process of minimizing informational entropy*.

## 4. The Energy Model of a Formal Neuron

The McCulloch-Pitts (MCP) neuron model is a simplified mathematical representation of a biological neuron, which, nevertheless, has become widely used and has formed the basis for the construction of modern ANNs. This informational model is implemented algorithmically; however, modeling it as an energy system will allow us to understand the limitations of this model compared to a real physical system, which a biological neuron undoubtedly is.

Let us consider the MCP model as a single-layer perceptron in its classical version with binary signals {0, 1}, weight coefficients in the range {−1, +1}, and a Heaviside activation function. The energy model of the MCP can be examined based on the construction of its energy landscape. We have already mentioned that the energy landscape of an MCP neuron in its resting state is represented by a set of weight coefficients that determine the energy contribution of each input to overcoming the "potential energy well," the depth of which is limited above by the activation threshold $Q$ (12-16).

Additionally, we have noted that the energy of the MCP model represents only the energy of a single unit in the energy landscape of a more complex system (e.g., an ANN). This unit (neuron) does not form energy interactions with other units, unlike real physical systems (9).

Figure 2 shows the general structure of the MCP energy model, where $X = (x_1, x_2, ...,$ $x_n)$ is the set of input signals representing the system's input energy; $Y$ is the response of the MCP model, representing the system's output energy; $W = (w_1, w_2, ..., w_k)$ are the weight coefficients representing the energy contributions of the inputs to overcoming the activation threshold; $\Sigma$ is the summator of the weighted inputs, representing the current depth of the system's (unit's) "potential well"; and $f$ is the Heaviside activation function, representing the upper threshold of the "potential well."

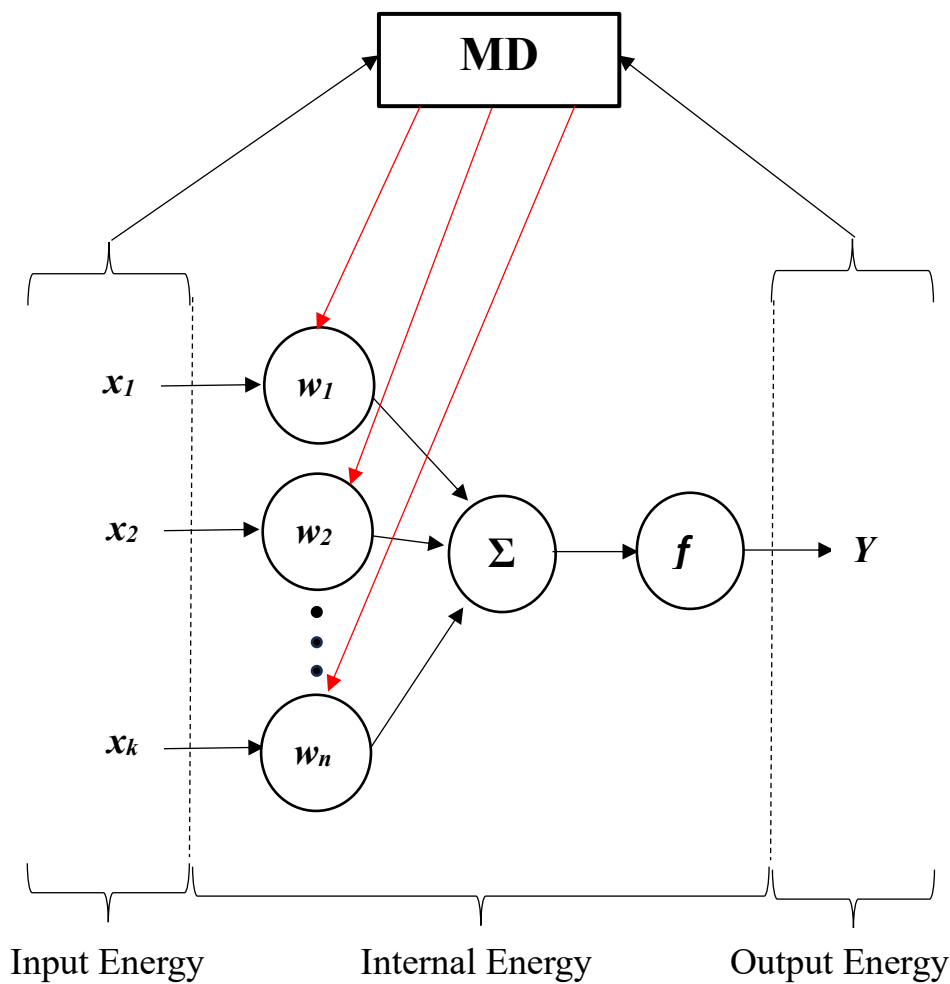

Figure 2. General structure of the energy model of the MCP.

The sum of the weighted inputs represents the internal energy of the system. This energy is divided into excitatory (activation energy), where the weight coefficients have a positive sign, and inhibitory (suppressing) energy, where the weight coefficients have a negative sign.

Thus, if the "positive" total input energy exceeds the "negative" or inhibitory energy, the activation threshold $Q = 0$ is surpassed, and the system, using the activation function, decides to generate an output signal $Y = 1$, regardless of the activation potential $V$:

$$V = \sum_i w_i^+ x_i + \sum_i w_i^- x_i = V^+ + V^- \qquad (23),$$

where $w_i^+$ are the positive weight coefficients that determine the positive activation potential $V^+$, $w_i^-$ are the negative weight coefficients that determine the inhibitory potential $V^-$.

Thus, the inhibitory potential $V^-_{rest}$ determines the "depth" of the potential well and the minimum energy of the unit in the resting state $V_{rest}$:

$$V_{rest} = V^-_{rest} = \sum_i w_i^- \qquad (24).$$

This "depth" of the potential well is represented in the energy landscape of the unit model (formal neuron) in the resting state (Figure 3).

The process of normalizing the output energy of the system (representing it as a single-bit value based on Landauer's principle) can be associated both with the "dissipation" of energy exceeding the normalized value of "1" and with the use of additional energy in cases where $V < 1$.

If $|V^-| > |V^+|$, then all the energy $V^+$ will be dissipated with energy release in accordance with Landauer's principle (7).

The model in Figure 2 also presents "Maxwell's Demon" (MD) [21]. This is an abstract thermodynamic concept proposed by the Scottish physicist James Clerk Maxwell, which is widely used in information physics and quantum computing. In this case, we consider it as an analogy to an external energy system relative to the given system, which implements the learning algorithm of the neuron. As an energy system, MD regulates the flow of input energy for its transformation into the internal energy of the system by purposefully adjusting the weight coefficients.

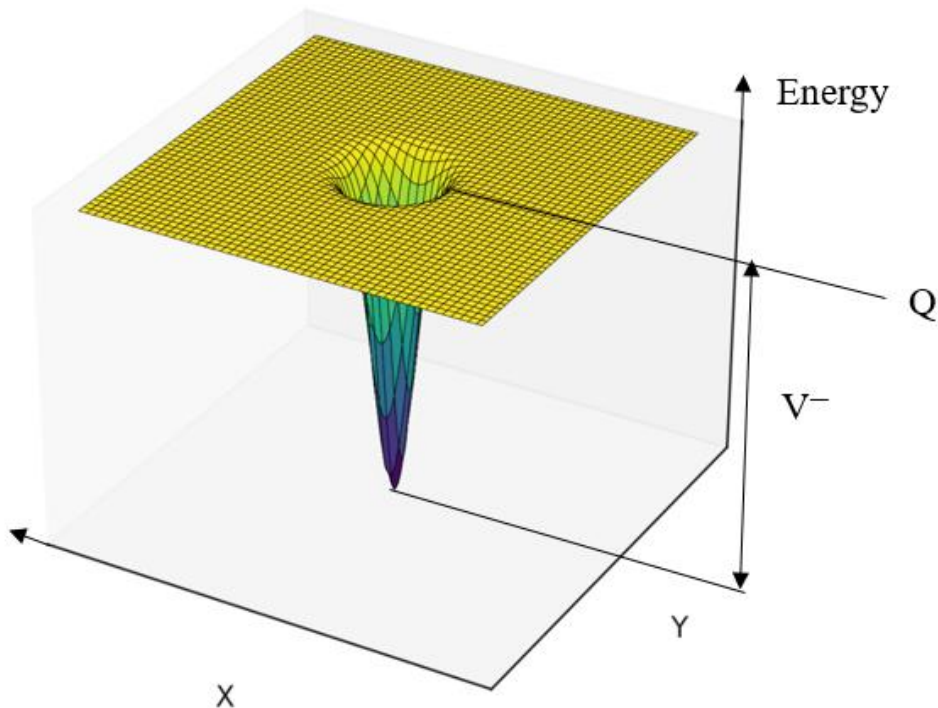

Figure 3. Energy landscape of the MCP model in the resting state.

Practically, MD minimizes the error energy at the system's output. This error energy represents "pseudo-energy" relative to the system's internal and external energy.

To consider the MCP model as a thermodynamic system, we must also define the system's microstates and macrostates. We define the system's microstates as the set of weighted inputs, which correspond to specific system outputs—macrostates. In this model, there are only two macrostates: $Y = \{0,1 \}$, where only the macrostate $Y = 1$ corresponds to neuron activation.

The probability of neuron activation $P_{act}$ can be considered as a conditional probability:

$$P_{act} = \sum_{V^-} P(V^-) P(V^+ \geq (Q + V^-)|V^-) \qquad (25),$$

where:

$P(V^-)$ is the probability that the inhibitory inputs sum to $V^-$;

$P(V^+ \geq (Q + V^-)|V^-)$ is the conditional probability that the excitatory inputs sum to a value exceeding the effective threshold $(Q + V^-)$.

This reflects the influence of inhibitory contributions: the "deeper" (i.e., the larger in magnitude $V^-$ is), the higher the effective threshold for excitatory inputs, and thus, the lower the probability that the condition $V^+ \geq (Q + V^-)$ will be satisfied.

If we define the number of microstates leading to activation as $\Omega_{act}$, then the thermodynamic entropy $S_{therm}$ is given by:

$$S_{therm} = k_B \ln \Omega_{act} \qquad (26).$$

The number of microstates $\Omega_{act}$ depends on how many combinations of weighted excitatory and inhibitory inputs satisfy the condition $V^+ \geq (Q + V^-)$. That is a "deeper well" (stronger inhibition) reduces $\Omega_{act}$ because more excitation is required to overcome the threshold, leading to a decrease in thermodynamic entropy for this activation macrostate.

Similar reasoning applies to the second macrostate $Y = 0$, where the number of microstates that do not lead to activation, $\Omega_{non-act}$, is determined by the condition $V^+ < (Q + V^-)$.

Assuming an equiprobable distribution of all $2^N$ microstates, where $N$ is the number of inputs, the probability of activation can be expressed as:

$$P_{act} = \frac{\Omega_{act}}{2^N} \qquad (27).$$

Then

$$\Omega_{act} = 2^N P_{act} \qquad (28).$$

In this case, the thermodynamic entropy for the activation macrostate $S_{act}$ is given by:

$$S_{act} = k_B \ln(2^N P_{act}) = k_B(N \ln 2 + \ln P_{act}) \qquad (29).$$

*Thus, as $P_{act}$ increases, the thermodynamic entropy also increases for a fixed value of $N$ and an equiprobable distribution of microstates.*

Similar reasoning applies to the determination of the thermodynamic entropy for the second macrostate $S_{non\text{-}act}$.

However, during the learning process of a neuron, we restrict the number of permissible input combinations, thereby making them non-uniformly distributed for a given (defined) macrostate. That is, for example, to activate a neuron, we apply the conditional probability given by equation (25). In this case, we use only a significant subset $T$ from the total set of $2^N$ combinations:

$$|T| \ll 2^N \qquad (30).$$

If we assume that the distribution of combinations within $T$ is uniform, then:

$$P_{act} = \frac{\Omega_{act}}{|T|} \qquad (31),$$

which means that maintaining the same probability $P_{act}$ under conditions of a non-uniform distribution of input combinations—leading to the selection of a significant subset $T$—requires a smaller number of effectively active input combinations and, consequently, a smaller number of possible microstates $\Omega_{act}$. The same reasoning applies to the subset of combinations $T'$ that do not lead to neuron activation.

*Thus, we observe that during the learning process of a neuron, its thermodynamic entropy as an energy system decreases for the selected significant subsets of input combinations T and T'. In contrast, for the overall equiprobable distribution of input combinations, the thermodynamic entropy for both macrostates increases as the number of microstates grows to $2^N$.*

Let us consider the informational (Shannon) entropy $H$ of the MCP model:

$$H = -[P_{act} \log_2 P_{act} + (1 - P_{act}) \log_2 (1 - P_{act})] \qquad (32).$$

We see that $H$ reaches its maximum at $P_{act} = 0.5$, i.e., when the distribution of weighted excitatory and inhibitory inputs is equiprobable. This also corresponds to the *maximum thermodynamic entropy*.

*When the conditional probability $P_{act}$ (equation 25) approaches 0 or 1, we observe a decrease in informational entropy, which also corresponds to a local decrease in thermodynamic entropy for the corresponding macrostate.*

Thus, if we define informational entropy through the probability distribution over microstates, it turns out that it precisely corresponds to thermodynamic entropy (considering the constant $k_B$). Moreover, the learning process, which leads to the selection of a significant subset of input combinations, modifies the distribution of microstates, thereby causing a local reduction in thermodynamic entropy.

Now, let us consider the output energy of the system as the Gibbs free energy $G$ [22, 23]:

$$G = H - TS \qquad (33),$$

where $H$ is the enthalpy of the system (heat content, energy including heat), $T$ is the absolute temperature, and $S$ is the thermodynamic entropy of the system.

Gibbs free energy is a thermodynamic state function that characterizes the system's ability to perform useful work under isothermal and isobaric conditions (i.e., at constant temperature and pressure). This term is widely used in chemical thermodynamics to analyze the energetic aspects of chemical reactions and phase transitions. However, its interpretation as the output energy of a system is not commonly used. Generally, it indicates how much energy can be converted into useful work—mechanical, electrical, or other forms.

Importantly, Gibbs free energy can be linked to the internal structures of a system (e.g., crystal lattice, molecular organization). In this case, the value of $G$ reflects how the system's structure influences its ability to perform work. Nevertheless, in machine learning and Bayesian statistics, the method of variational free energy is applied, utilizing thermodynamic principles for optimization [24]. Here, a functional analogous to Gibbs energy is introduced to minimize the difference between the model and the data.

However, if we consider that the free energy at the output of an information system "performs work" by creating an internal model of the external world for another system that perceives it, then the analogy between information generated at the system's output and Gibbs free energy takes on a completely different meaning and becomes well-justified. Furthermore, information enables systems to make decisions, predict, or change their state. For example, in control systems or robotics, the output of an ANN is used to control real physical objects, which serves as an analog to physical work.

Then,

$$G = E_{str} = E_{total} - E_{unstr} \qquad (34),$$

where $E_{str}$ is the structured part of the energy at the system's output. This portion of energy contains measurable parameters that reflect the order associated with the internal structure of the system. It is related to the reduction of informational entropy due to data structuring (e.g., classification, feature extraction, prediction, etc.);

$E_{total}$ is the total energy released by the system (enthalpy);

$E_{unstr}$ is the unstructured part of the energy. This portion of energy remains in a form that does not carry a stable structure in the system's output energy, which could otherwise represent information about the system's internal structure. This energy is analogous to thermal losses or noise, which increase thermodynamic and informational entropy.

We observe that in the MCP model, only for the macrostate "1" does the formation (release) of free energy occur, whereas for the macrostate "0," which signifies the absence of a neuronal response, all energy represents thermodynamic losses (*TS*) in accordance with Landauer's principle. If we disregard the energy costs associated with normalizing the unitary output of the neuron—additional energy for $0 \leq V < 1$, or additional heat dissipation for $V > 1$—then it can be assumed that the inhibitory potential $V^-$ makes a significant contribution to the formation of $TS = E_{unstr}$

$$TS = \Omega V^- E \qquad (35),$$

where $E$ is Landauer's energy (7), defining the thermal energy released during the erasure of one bit of information; $\Omega$ is the total number of microstates; $V^-$ is the inhibitory potential, determining the number of bits erased per microstate.

Then, the total unstructured portion of the emitted energy, as the "chaotic" part of enthalpy, is equal to the total thermodynamic entropy of the MCP neuron:

$$S = \Omega V^- k_B \ln 2 = 2^N V^- k_B \ln 2 \qquad (36).$$

We observe that the total thermodynamic entropy of the MCP model increases with the number of microstates and the "depth" of the neuron's potential well (inhibitory potential).

Consequently, as the total number of microstates in the system increases, the system's enthalpy grows, and its unstructured component $E_{unstr}$ also increases, while the structured part of the output energy $E_{str}$ reaches its maximum value for a limited number of microstates after learning. This corresponds to the minimum thermodynamic and informational entropy for the respective macrostates and remains unchanged thereafter.

This conclusion aligns with the findings of Prigogine's theory of dissipative systems, which states that a system, when far from equilibrium, can "learn," meaning it transitions into a state that acts as an attractor in its dynamics. This implies that out of a vast number of possible microstates, the system "selects" those that optimally maintain order (low internal entropy) through energy dissipation. Thus, although the total entropy of the environment may continue to increase, the internal structure of the system stabilizes.

Let's clarify that in classical equilibrium thermodynamics, Gibbs free energy is minimized. However, here we are dealing with an open system which, considering "Maxwell's Demon," can only conditionally be regarded as a self-organizing system. In this case, the concept of the "maximum" of the structured energy $E_{str}$ should be understood in terms of the system's optimal functional state rather than as a classical equilibrium state.

This does not contradict the classical interpretation of Gibbs free energy, since we observe that as the system's total enthalpy increases, its total thermodynamic entropy $E_{unstr}$ also increases, whereas the free energy $E_{str}$ —or more precisely, its fraction as a component of enthalpy—decreases, along with its local thermodynamic entropy.

Overall, these conclusions are consistent with *the system's dynamic evolution equation*, which is analogous to the Langevin equation in statistical mechanics, describing the motion of a particle in a potential field with thermal noise

$$\frac{dx}{dt} = -\frac{\partial V(x)}{\partial x} + \eta(t) \qquad (37),$$

where $x$ represents the internal energy of the system, $V(x)$ is the effective energy potential landscape, and $\eta(t)$ is a stochastic noise term modeling thermal fluctuations associated with thermodynamic losses [25, 26].

In this equation, the function $V(x)$ serves as an effective energy potential, determining how the system tends to minimize its internal energy. Thus, $-\partial V(x)/\partial(x)$ defines the natural tendency of the system to reduce its energy, i.e., to transition to more stable states.

If a dissipation coefficient $\gamma$ is added to the left-hand side of the equation, we obtain an equation for a dissipative system, which resembles the relaxation equation for energy in open systems.

*General Conclusions*:

1. In the MCP model, as well as in ANN models, information is represented as abstract elements of the system's internal alphabet, such as bits. These alphabet elements lack an internal structure that could be associated with energy structures, as seen in quantum physics or chemistry. However, informational bits are linked to energy according to Landauer's principle, allowing for an energetic model to be considered for both neurons and ANNs as a whole. In this case, the internal energy of the MCP model, as a unit of the energy system, can only be analyzed based on external "pseudo-energy," whose parameters (weight coefficients) are determined by "Maxwell's Demon" (an external training algorithm) through statistical observations. The statistical parameters of external "pseudo-energy" for determining the internal energy of a unit (MCP model) can be regarded only as "surface" properties of the system's energy, which do not fully reflect the structure of the unit's internal energy and cannot be used to form the energy of unit interactions, as occurs in physical systems.

2. During the training process of the MCP model, a stable, structured component of the energy emitted (generated) by the system is formed, which can be considered as Gibbs free energy. This structured energy exhibits a minimal thermodynamic and informational entropy for a limited number of effective microstates corresponding to certain (predefined during training) macrostates. The decrease in both entropies follows the same dynamics, reflecting the deep energetic nature of their interrelation. During training, the MCP model—more precisely, "Maxwell's Demon" as part of it—minimizes the external "pseudo-energy" in the form of the model's response error. This minimization is manifested in changes to the statistical parameters of the unit's internal energy, represented as weight coefficients. This leads to a reduction in the number of effective microstates and, ultimately, to the minimization of the system's internal energy.

3. Based on the presented interpretation of free energy, it can be stated that the parameters of structured energy at the system's output (for the MCP model, this could be a sequence of binary signals) reflect the internal energy of the system and, consequently, its internal structure, which can be interpreted as an *internal model of world perception*.

## 5. Energy Models of Multilayer ANNs

To understand the energetic nature of information, let us consider multilayer ANNs, which can be regarded as ideal models not only for studying the principles of information processing but also for examining their relationship with the energy of physical systems. To this end, we will use the concept of an energy landscape and the energy model of the MCP previously discussed.

Figure 4 presents the energy landscape model of a multilayer perceptron (MLP).

This energy landscape consists of potential "wells" of units (MCP neurons) of varying depths in the resting state, forming the hidden layers. The model illustrates the general concept of energy convergence, which reflects the process of information convergence in an MLP when solving a classification task.

A key feature of this landscape is the absence of energy connections—channels for energy distribution between units. The output energy of units simply dissipates among the inputs of the next layer's units.

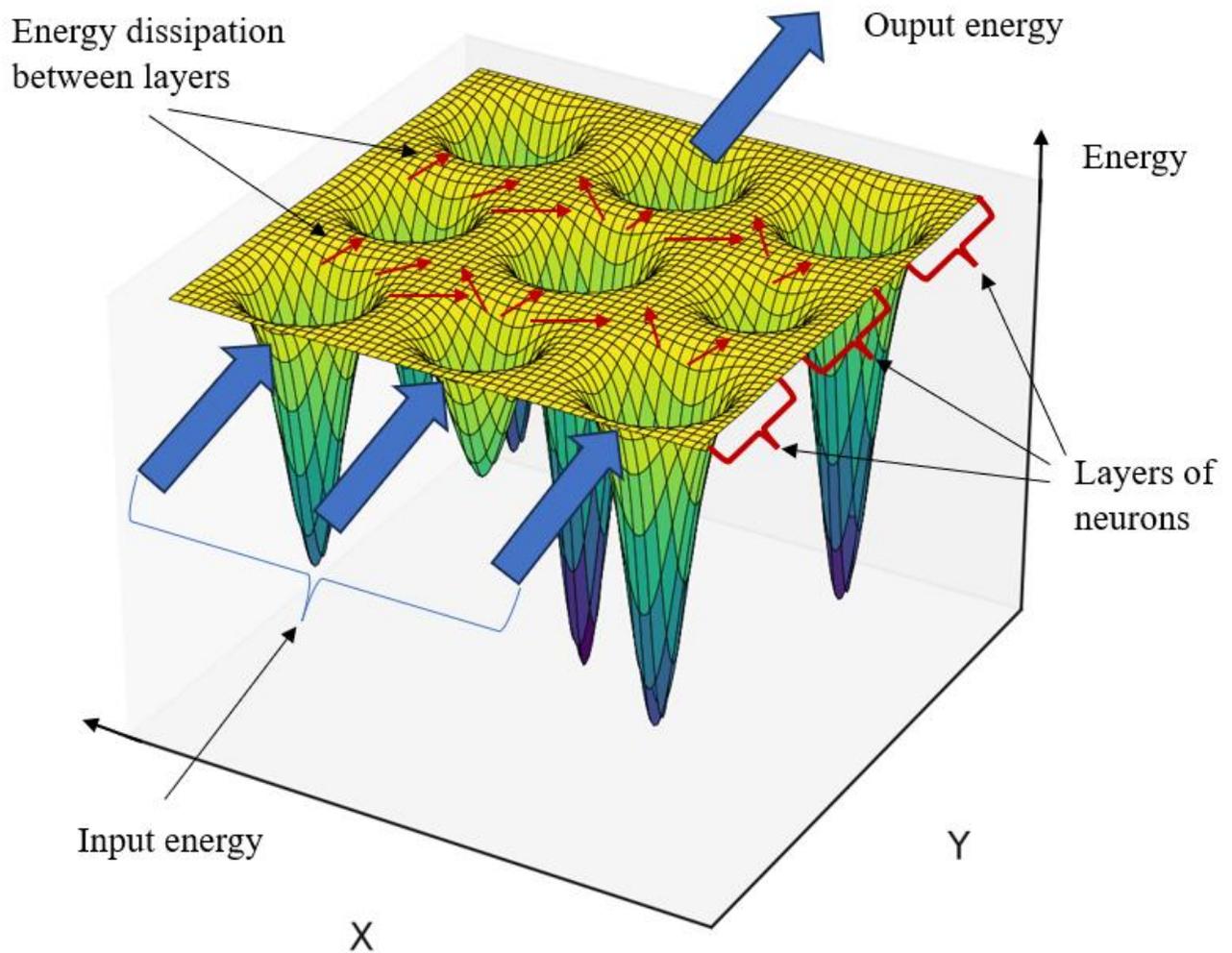

Figure 4. The energy landscape model of MLP.

This results from the fact that the input weight coefficients do not generate connection energy but merely determine the depth of a unit's potential "well." The process of output energy dissipation is represented by the fully connected architecture of MLP layers. In a real energy system, such dissipation would require amplification of the information signal at each neuron input. The lack of connection energy between structural elements in this energy model is a limitation compared to real energy systems (Figure 1).

The coordinated formation of unit (neuron) energy in this ANN model (i.e., the formation of the system's energy landscape in its resting state) is also linked to the process of minimizing "pseudo-energy," governed by the system's "Maxwell's Demon" (e.g., the backpropagation algorithm). This process of coordinated unit energy formation, aimed at achieving the required set of macrostates, represents a trade-off between the generality of the extracted classification features and the accuracy of the classification task for a given set of classes. In many practical tasks, achieving this trade-off during training does not allow for a perfect solution.

Based on the research results presented in Section 4, it can be concluded that the informational and thermodynamic entropies of a system whose energy landscape consists of multiple separate energy units (MCP models) will exhibit the same properties as an individual unit.

Evidence for these conclusions comes from studies by Ravid Shwartz-Ziv and Naftali Tishby (2017) et al. [27, 28], in which the authors proposed the Information Bottleneck (IB) theory. This theory examines deep neural networks, such as MLPs, and demonstrates that during training, there is a tendency for information entropy to decrease in the responses of hidden-layer neurons. Moreover, the reduction in information entropy is accompanied by a decrease in the variability of neuron responses in hidden layers, indicating a reduction in the number of microstates for the considered responses (macrostates) of the system. This also suggests a local decrease in thermodynamic entropy for these macrostates.

For example, in the work of Shwartz-Ziv and Tishby (2017) [27], it is shown that in classification tasks on datasets such as MNIST, the information entropy of activation decreases by a factor of 2–4 from the input layer to the output layer, corresponding to a reduction in activation variance from 50% to 90% between the initial and deeper layers.

In IB theory, the authors introduce the concept of mutual information: $I(X; T)$ – the information that layer $T$ contains about the input data $X$; $I(Y; T)$ – the information that layer $T$ contains about the target data $Y$.

Mutual information $I(X; Y)$ determines how much knowing one variable reduces the uncertainty of the other. In general, mutual information equals the sum of the entropies of the individual variables minus their joint entropy

$$I(X;Y) = H(X) - H(X|Y) \tag{38},$$

$$I(X;\text{Y}) = H(Y) - H(Y|X) \tag{39},$$

$$I(X;Y) = I(Y;X) \tag{40},$$

$$I(X;Y) = H(X) + H(\text{Y}) - H(X;Y) \tag{41}.$$

If $X$ and $Y$ are independent, knowing $Y$ does not reduce the uncertainty of $X$, then:

$$H(X|Y) = H(X) \ and \ I(X;Y) = 0 \tag{42}.$$

If $X$ and $Y$ are fully dependent, knowing one variable completely determines the other, and $I(X; Y)$ reaches its maximum value, equal to the entropy of the variable with the smaller entropy.

These studies show that $I(X; T)$ decreases significantly toward the final layers, indicating a substantial reduction in the uncertainty of the input data. Moreover, the information entropy of activation $H(T)$ in each layer decreases during training. For example, in a 10-layer network, $H(T)$ in the last layer decreases by approximately 50-70% compared to the first layer. However, as expected, information entropy remains at the system's output, confirming the existence of a trade-off between the generality of extracted classification features—determined by the system's generalized energy landscape—and the stability (accuracy) of the system's responses to specific input information.

It has also been shown that, at the beginning of training, mutual information between a layer's representation and the input $I(T; X)$ in the initial layers can reach about 10–15 bits, meaning that a large amount of raw input information is preserved at these layers. As information propagates through deeper layers, $I(T; X)$ begins to decrease. For example, in the second hidden layer, $I(T; X)$ may drop to 8–10 bits. The most significant information compression occurs in the last hidden layers, where $I(T; X)$ can be reduced to 2–3 bits, while $I(T; Y)$ remains close to the entropy of the labels. This indicates that all relevant informational features necessary for decision-making are retained.

These empirical results fully confirm our theoretical conclusions about the reduction of information entropy and the local decrease in the structured part of thermodynamic entropy at the output of the considered system (the energy model of an artificial neural network) for a specific set of its macrostates.

Thus, since the considered system lacks energy connections between units, its internal energy is represented exclusively by the energy of the units themselves. Each unit perceives the energy of a neighboring unit as external energy, converting it into its internal energy using weight coefficients, which act as specific detectors of input energy. Consequently, the activation function of a neuron can be interpreted as a function that transforms internal energy into the output energy of the unit.

The structured external energy of the entire system, which represents the information at the output of the ANN, is formed by the activation functions of the neurons in the outer (final) layer. This system implements a process of parallel energy (information) perception and its convergence (compression) during training.

*Let us call such an energy model of an ANN a first-type model*. This type also includes convolutional neural networks (CNNs), whose main distinction lies in the segmentation of the input signal vector (input energy) and the local processing of individual segments.

The first-type ANN model can be classified as an energy system with: *a deterministic static structure, a deterministic static generalized energy landscape in the resting state, iteratively formed by the system's "Maxwell's Demon," a deterministic energy dynamic of units on the energy landscape, occurring during the minimization of informational entropy and the local minimization of thermodynamic entropy*.

We can also conclude that the generalized energy landscape of the system in the resting state, represented solely by the energy of units, along with the deterministic static structure of the system, imposes a conceptual limitation on its accuracy as the specified number of macro-states of the system increases.

In text-generative ANN models such as GPT, unlike first-type ANN models, there is a divergence between input energy and information, which we associate with "predictions." These systems have an entirely different energy landscape, which is not represented by the energy of units, as in MLP models, but rather by the energy of connections between units, while the units themselves function only as static elements of the system's internal alphabet.

For generative ANN models, these elements are, for example, words of a natural language (or their components), represented as tokens and their internal numerical representations - embeddings.

In general, static embeddings, such as those used in Word2Vec or GloVe, represent words as fixed-length vectors in a high-dimensional space. The proximity of these vectors (e.g., measured by cosine similarity) can serve as an indirect indicator of the similarity or association between words. This associative connection can be conditionally interpreted as a "probability of relation" between words, although this is not a direct probabilistic measure in the mathematical sense. Instead, it reflects only the averaged statistical relationships from the training corpus. Such a vector can be considered a representation of the "central meaning" of a word, but it does not account for polysemy or meaning shifts across different contexts.

In contrast, contextual embeddings in transformer-based models, such as BERT or GPT, dynamically transform a word's base representation—a lookup vector—based on surrounding words (context). This process highlights those aspects of the word's meaning that are relevant in the given context. The same word may have different vector representations depending on syntax, semantics, and even the text's style, allowing the model to account for polysemy.

This mechanism enables the model to dynamically select—or rather compute—the most relevant (most probable, given a Softmax activation function) word relationships for a given context, based on the query-key-value attention matrix. The dynamics of changing relationships during inference are implemented through the Multi-Head Attention mechanisms across different layers of the model.

All word (embedding) relationships are formed based on weight coefficient matrices, which in turn are learned during training (in the energy interpretation—by "Maxwell's Demon"). This means that all potential pathways of word relationships are "embedded" in the model's weight matrices, as these matrices define the structure of possible associations and transformations. However, the specific relationships relevant to a given context are selected (computed) dynamically when processing input text.

Thus, the model's weight matrices establish the depth of "potential channels" of relationships between units in the system's energy landscape in its resting state. The activation functions of the model's neurons determine the "output energy" from these channels, which in turn activates the "static energies" of the units. These unit energies collectively form the structured part of the system's output energy.

In this sense, the neurons of generative models effectively establish relationships between the elements of the system's internal alphabet (units).

A simplified model of the energy landscape of a generative ANN is shown in Figure 5.

Thus, this system implements the process of sequential perception of the energy (information) flow and its divergence across various "channels" of the system's energy landscape to activate the conditional energy of units (tokens).

*Let us call such an energy model of an ANN a second-type model*. This type may include recurrent ANNs as well as generative models based on transformers.

This model has a complex energy landscape formed by numerous weight matrices of various types (in transformer models, about 8 types of matrices are used for attention mechanisms, embeddings, linear transformations in fully connected layers, normalization, and output formation). It also consists of multiple hidden layers and neurons in each layer (for example, in the GPT-3 model, the hidden layer size can be up to 12,288 neurons for the largest model with 96 layers), which determines an enormous number of model parameters (GPT-3 contains up to 175 billion parameters). These parameters and their combinations form a vast space of possible microstates of the system.

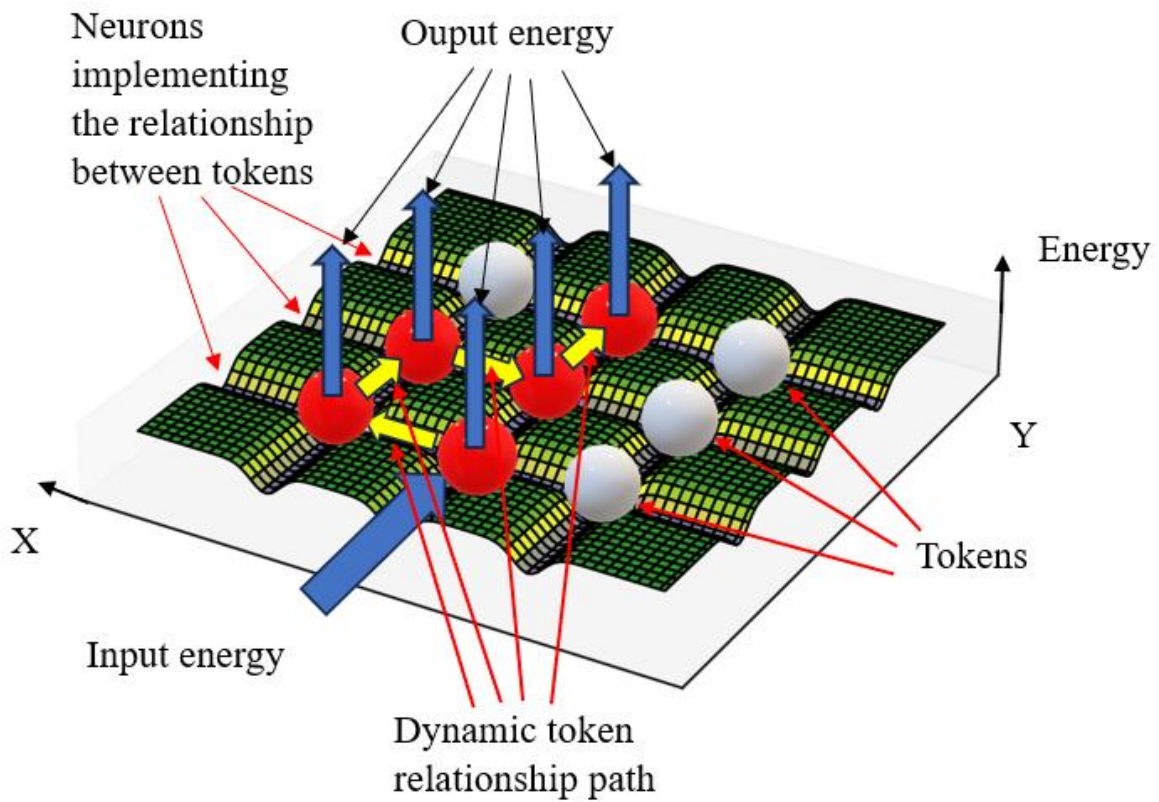

Figure 5. The energy landscape model of generative ANN.

Then, the dynamics of selecting the most relevant token (unit) connections during inference (the operation of a trained generative ANN) determine the reduction of the microstate space to its minimal possible extent (depending on the token selection mechanism). Ideally, it collapses to a single fixed set of microstates, which then defines the system's macrostate in the form of output information.

Thus, during training, the generative ANN model expands the space of possible system microstates (forms it), while during inference, it reduces this space to a minimum, which determines the system's macrostates. At the same time, a first-type model reduces the microstate space during training for a given macrostate.

It is evident that the information entropy at the output of a second-type ANN also decreases. This is because generative models select the most relevant connections during inference—that is, the most probable contextual structure of tokens. This process reduces the number of possible token combinations, thereby decreasing the information entropy of the generated text, ideally collapsing it to a single structure for a given model.

However, in practice, heuristic algorithms such as Beam Search may be used to maximize the probability of a token sequence by selecting several of the most likely options. This leads to more predictable but less diverse results. This behavior can be seen as an analogy to the bottleneck effect in first-type ANNs.

To increase the diversity of texts (responses) in generative models (e.g., GPT-4), information entropy at the system's output is artificially increased. This is achieved by applying stochastic methods such as temperature scaling, which adds randomness to token selection, or Top-k/Top-p sampling, which limits the candidate pool for the next token selection, introducing additional randomness and enabling the generation of more diverse texts. Because of this, the space of microstates does not collapse into a single outcome but remains probabilistic, allowing for variation in model responses.

Thus, the statistical nature of generative ANN training leads to the output of the most probable, i.e., contextually averaged, token structure. This naturally results in reduced information entropy at the system's output. However, as discussed earlier, the most informative and potentially valuable connection for problem-solving is the one with low probability, as it represents new or "creative" information.

We also see that reducing the number of possible system microstates during inference, leading to the formation of a single macrostate at the output, indicates a local reduction in the structured component of the system's thermodynamic entropy (34).

Thus, a second-type ANN model can be classified as an energy system with: *a deterministic static structure, a deterministic generalized energy landscape in a resting state, iteratively formed by "Maxwell's Demon", and a deterministic dynamic change in unit connection energy on the energy landscape in the process of minimizing information entropy and locally minimizing thermodynamic entropy*.

*Conclusions*. The results of these studies indicate the conceptual limitations of first-type and second-type ANN models. The structure of the internal energy of these systems in a resting state is formed by weight coefficients, which are obtained by another system— a "Maxwell's Demon"—that is external to the considered systems. These weight coefficients are formed based on statistical observations and are in no way connected to external energy (information). These limitations prevent the realization of a self-organization process in the system based on energy, as occurs in real physical systems, such as quantum systems.

## 6. Information

### 6.1. Evolution of the Energy Landscape. Structural and Parametric Reduction

Let's define the total internal energy of the system $E_{total}$ as the sum of the internal energies of its structural elements (units) and the interactions between them:

$$E_{total} = \sum_i E_i + \sum_{i<j} E_{ij} \tag{43},$$

where $E_i$ is the internal energy of structural element $i$, and $E_{ij}$ is the energy of the interaction between elements $i$ and $j$.

Let's denote the number of true structural events of the input energy $N_{true}$ that are valid for the given macrostate of the system. In this case, by truth we mean the correspondence of the input energy structure to the specified macrostate of the system. In modern ANNs, true events refer to examples from the training set. Then, $N_{total}$ is the total number of events perceived by the system. We can define the weight coefficient www of true events that affect the system's energy:

$$w = \frac{N_{true}}{N_{total}} \tag{44}.$$

We will denote this dependency as:

$$E_i = E_i(w), \quad E_{ij} = E_{ij}(w) \tag{45}.$$

Next, let's define how $E_{total}$ changes as the number of true events $N_{true}$ increases:

$$\frac{dE_{total}}{dN_{true}} = \sum_i \frac{\partial E_i}{\partial w} \cdot \frac{\partial w}{\partial N_{true}} + \sum_{i<j} \frac{\partial E_{ij}}{\partial w} \cdot \frac{\partial w}{\partial N_{true}} \tag{46}.$$

The weight coefficient $w$ increases as the number of true events $N_{true}$ increases. To demonstrate this process, let's find its derivative with respect to $N_{true}$:

$$\frac{dw}{dN_{true}} = \frac{d}{dN_{true}}\left(\frac{N_{true}}{N_{total}}\right) \tag{47}.$$

If the total number of events $N_{total}$ is fixed (e.g., limited by the size of the training set), then the derivative simplifies to:

$$\frac{dw}{dN_{true}} = \frac{1}{N_{total}} \tag{48}.$$

Thus:

$$\frac{dE_{total}}{dN_{true}} = \frac{1}{N_{total}}\left(\sum_i \frac{\partial E_i}{\partial w} + \sum_{i<j} \frac{\partial E_{ij}}{\partial w}\right) \tag{49}.$$

Under the conditions:

$$\sum_i \frac{\partial E_i}{\partial w} < 0 \; and \; \frac{\partial E_{ij}}{\partial w} < 0 \tag{50},$$

the energy function of the system $E_{total}$ also decreases:

$$\frac{dE_{total}}{dw} = \sum_i \frac{\partial E_i}{\partial w} + \sum_{i<j} \frac{\partial E_{ij}}{\partial w} < 0 \tag{51}.$$

This condition is met only if, with an increase in the statistical significance of structural elements or connections, their energy parameters transition to more stable, invariant values. This means that their internal energy must have a structure, for example, in the form of discrete energy levels. In this case, an increase in the weighting coefficient determines the parametric commonality of energy, i.e., a transition to a general, more energetically stable level. This process of energy reduction can be compared to the transition from quantitative values of individual energy parameters to more general and invariant qualitative parameter values.

Thus, the growth of $w$ can be interpreted as a process of system ordering, in which an increase in the statistical significance of "true" events leads to changes in internal parameters so that the system transitions from a less ordered (high-energy) state to a more ordered (low-energy) state. Mathematically, this is reflected in the negative derivative $dE_{total}/dw$ (51), and physically – in a phase transition to a state with minimal energy.

A similar process of energy reduction also occurs in quantum systems, for example, when electrons transition to lower-energy orbitals.

*Thus, relations (46-50) define the evolution of the system's energy landscape (its structure and energy) during learning, while conditions (51) determine the interdependence of the energy of units and their interactions, i.e., the process of forming the system's internal energy. These conditions form the basis for constructing the rules of structural and energetic (parametric) reduction of the system.*

*The stage of structural reduction* is associated with the decrease in the number of significant (statistically relevant) structural elements and their connections. The reduction of the structure to statistical significance directly leads to a decrease in the system's total energy.

To illustrate this process, let us introduce a threshold of statistical significance for the weighting coefficients of units $w_i$ and their connections $w_{ij}$: $w_i \geq \gamma$, $w_{ij} \geq \gamma$, as well as the structural reduction operator $R_w$, which excludes statistically less significant structural components, leading to the formation of a reduced structure with energy

$$E_{red} = R_w(E_{total}) < E_{total} \qquad (52).$$

*The stage of parametric reduction* of the system's internal energy is associated with the reduction of the energy of units and their interactions.

To satisfy this condition, the internal energies of the units $E_i$ and their interconnections $E_{ij}$ *must be interdependent*.

Let us define:

$U^{(0)} = \{u_i^{(0)}\}$ as the set of initial quantitative variables—measurable parameters that characterize the energy state of the units and determine the structure of their energy.

$C^{(0)} = \{c_{ij}^{(0)}\}$ as the set of initial computed parameters (e.g., based on *the function comparing the measurable energy parameters of the units*), which characterize the energy state of the unit connections and determine their energy structure.

In the general case:

$$c_{ij} = \Phi\big(u_i, u_j\big) \tag{53},$$

where $\Phi$ is a comparison function that defines the interconnection parameter $c_{ij}$ as the result of comparing, in this case, the parameters of two units. Thus, equation (53) links the energy of the units with the energy of their interconnections. Moreover, this function forms the internal energy landscape of the unit interconnections—"channels" within the overall energy landscape of the system (Figure 1).

This occurs due to the discretization of energy levels when transitioning from quantitative to qualitative values of comparison parameters, i.e., when moving from a quantitative measurement scale to a qualitative scale. In general, there are two types of qualitative scales: nominal and ordinal, and two types of quantitative scales: interval and ratio scales, which are metric [29].

The "strongest" scale is the ratio scale, as it forms an absolute metric for the quantitative values of the parameters. That is, we can interpret its "strength" as the highest energy level. In contrast, the nominal qualitative scale merely categorizes features without quantitative measurement or ordering. This is the "weakest" scale, which we can associate with the lowest energy level.

Thus, the most invariant to quantitative parameter values and, at the same time, the most informative qualitative scale for defining interconnection energy is the ordinal scale. Metaphorically, its energy can be compared to the energy of an atomic $p$ orbital, which, due to its directionality and role in bond formation, plays a central role in atomic interactions [30].

At each structural level $L = 0, 1, 2, \ldots, L_{max}$, the system is characterized by:

- A set of unit parameters $U^{(L)} = \left\{ u_i^{(L)} \right\}$,

- A set of interconnection parameters $C^{(L)} = \left\{ c_{ij}^{(L)} \right\}$.

At level $L$, the total energy of the system $E_{total}$ is given by:

$$E_{total}^{(L)} = E_{units}^{(L)}\left(U^{(L)}\right) + E_{int}^{(L)}\left(U^{(L)}, C^{(L)}\right) \tag{54},$$

where

$$E_{units}^{(L)}\left(U^{(L)}\right) = \sum_i E_i^{(L)}\left(u_i^{(L)}\right) \tag{55},$$

$$E_{int}^{(L)}\left(U^{(L)}, C^{(L)}\right) = \sum_{ij} E_{ij}^{(L)}\left(c_{ij}^{(L)}\right) = \sum_{ij} E_{ij}^{(L)}\left(\Phi\left(u_i^{(L-1)}, u_j^{(L-1)}\right)\right) \tag{56}.$$

When transitioning from level $L$ to level $L+1$, parameter reduction is performed—moving from initial quantitative characteristics to more invariant qualitative ones, which leads to a decrease in energy:

$$E_{total}^{(L+1)} < E_{total}^{(L)} \tag{57}.$$

The dynamics of parameter changes are described by gradient descent along the energy landscapes of individual units and their interconnections at each level in the process of evolutionary formation of the system's overall energy landscape, i.e., in the learning process. The iterative relations for units and their interconnections can be written as:

$$U^{(L+1)} = U^{(L)} - \eta^{(L)} \frac{\partial E_{units}^{(L)}}{\partial U^{(L)}} \qquad (58),$$

$$C^{(L+1)} = C^{(L)} - \eta^{(L)} \frac{\partial E_{int}^{(L)}}{\partial C^{(L)}} \qquad (59),$$

where $\eta^{(L)} > 0$ is the transition step between measurement scales.

This process can be compared to energy processes in classical mechanics, where forces in the system are related to the gradient of potential energy: $F = -\Delta U$ (where $U$ is the potential energy). This means that if the system finds itself in a state of higher energy, the acting forces will be directed in such a way as to bring the system to a state of lower energy. Thus, any fluctuations that increase energy are "repelled" by force fields, bringing the system back to the minimum potential energy.

The condition for energy minimization at each level requires that, in the limiting case, the equilibrium conditions are satisfied:

$$\frac{\partial E_{total}^{(L)}}{\partial u_i^{(L)}} = \frac{\partial E_{units}^{(L)}}{\partial u_i^{(L)}} + \frac{\partial E_{int}^{(L)}}{\partial u_i^{(L)}} = 0 \quad \forall i \qquad (60),$$

$$\frac{\partial E_{total}^{(L)}}{\partial c_{ij}^{(L)}} = \frac{\partial E_{int}^{(L)}}{\partial c_{ij}^{(L)}} = 0 \quad \forall(i,j) \qquad (61).$$

At the same time, the interdependence condition holds:

$$\frac{\partial^2 E_{int}^{(L)}}{\partial u_i^{(L)} \partial c_{ij}^{(L)}} \neq 0 \tag{62},$$

which ensures that changes in the interconnection parameters $c_{ij}^{(L)}$ affect the optimal distribution of $u_i^{(L)}$ and vice versa. In particular, if changes in connection parameters lead to a decrease in $E_{int}$, for example, due to the establishment of a more energetically "favorable" connection, then the equilibrium condition for units (60) forces the system to adjust $u_i$ so that $E_{units}$ also decreases.

A similar energy interdependence is observed in physical system models, such as the Ising model.

To simplify the analysis at each level, unit parameters can be expressed as a function of interconnection parameters, i.e., $U^{(L)} = U^{(L)}(C^{(L)})$. Then, *the effective energy* is defined as:

$$E_{eff}^{(L)}\left(C^{(L)}\right) = E_{units}^{(L)}\left(U^{(L)}\left(C^{(L)}\right)\right) + E_{int}^{(L)}\left(U^{(L)}\left(C^{(L)}\right), C^{(L)}\right) \tag{63}.$$

The concept of effective energy allows for the description of complex systems by simplifying their analysis, i.e., accounting for microscopic processes through their averaging in order to focus on the macroscopic behavior of the system [31-33].

The total differential of the effective energy with respect to the interconnection parameters is given by:

$$\frac{dE_{eff}^{(L)}}{dC^{(L)}} = \frac{\partial E_{int}^{(L)}}{\partial C^{(L)}} + \left[\frac{\partial E_{units}^{(L)}}{\partial U^{(L)}} + \frac{\partial E_{int}^{(L)}}{\partial U^{(L)}}\right]\frac{dU^{(L)}}{dC^{(L)}} \tag{64}.$$

Under equilibrium conditions, according to the minimization conditions for $U^{(L)}$ (60), the expression in square brackets equals zero. Then:

$$\frac{dE_{eff}^{(L)}}{dC^{(L)}} = \frac{\partial E_{int}^{(L)}}{\partial C^{(L)}} \qquad (65).$$

Thus, the reduction of effective energy occurs due to changes in the interconnection parameters. This relation illustrates that reducing the interconnection energy $E_{int}$ by modifying $C$ automatically leads to a change in the state $U$ in such a way that the unit energy $E_{units}$ also decreases to its optimal state as a result of the transition from quantitative to less energetic qualitative parameters.

Therefore,

$$\frac{dE_{eff}^{(L)}}{dC^{(L)}} < 0 \qquad (66).$$

Thus, at the stage of parametric reduction $R_{u,c}$, the total energy of the system also decreases:

$$E_{red} = R_{u,c}(E_{total}) < E_{total} \qquad (67).$$

## 6.2. Structural-Parametric Reduction

One of the key stages of reduction, which forms the structure-attractor (concept) representing the structure of the system's energy landscape in a resting state, is *structural-parametric reduction*.

This reduction allows for the identification of substructures with invariant interconnection parameters between elements (units) and simplifies the system by eliminating elements that do not significantly affect the parametric stability of the substructure as a whole. These elements are the least informative, enabling the substructure to be compressed down to two elements—two extreme or critical elements—critical structural points: the initial and final points of the substructure, where the interconnection between them retains the same general parameters as for all elements within the given substructure.

An analogy can be drawn between this process and data compression in information systems, while the critical points themselves can be compared to bifurcation points in the theory of dissipative or dynamic systems.

Let us examine this type of reduction. Suppose there exists a sequentially ordered structure $S$, consisting of elements $s_i$, where each element is characterized by a set of homogeneous parameters $u_i$:

$$S = (s_1, s_2, \ldots, s_n), \qquad s_i = \{u_{i1}, u_{i2}, \ldots, u_{im}\} \qquad (68).$$

In this sequence of structural elements, we select the first element as the initial critical point of a stable substructure that we aim to define. Our task is to find the final critical point and reduce the identified substructure.

We introduce a comparison operation $\Phi$ for the homogeneous parameters $u_k$ of two neighboring structural elements $s_i$ and $s_{i+1}$, which determines their interconnection $c_{i,i+1}$ similarly to formula (53):

$$c_{i,i+1} = \Phi\big(u_{i,k}, u_{i+1,k}\big) \qquad (69).$$

If the parameters are quantitative, the difference between their values is mapped onto an interval scale using the function $\varphi$:

$$\Phi\big(u_{i,k}, u_{i+1,k}\big) = u_{i,k} - u_{i+1,k} = \varphi(\Delta u_{i,i+1}^k) \qquad (70).$$

If, for the next structural pair of elements $u_{i+1,k}$ and $u_{i+2,k}$, ordered by index $i$, the difference in these parameters differs from the first pair, then to unify the interconnection parameters $c_{i,i+1}$ and $c_{i+1,i+2}$ to a single value $u_k$, we use the first derivative:

$$\frac{du_k}{ds_i} \approx u_{i,k} - u_{i+1,k} = u_{i+1,k} - u_{i+2,k} \tag{71}.$$

In this case, for finding an invariant value for both interconnections, we are not primarily interested in the derivative value itself but in its gradient, i.e., the sign of this difference:

$$sign\left(\frac{du_k}{ds_i}\right) = \begin{cases} +1, & parameter\ increase \\ 0, & parameter\ constancy \\ -1, & parameter\ decrease \end{cases} \tag{72},$$

Then, if the gradient signs match, we obtain:

$$c'_{i,i+1} = c'_{i+1,i+2} = \sigma\left(sign\left(\frac{du_k}{ds_i}\right)\right) \tag{73},$$

where $\sigma$ is a function mapping the gradient (the derivative of the qualitative parameter) onto an ordinal qualitative scale.

For a structure consisting of $n$ elements, we can consider the vector of qualitative relationship parameters for all pairs of adjacent elements:

$$C'_S = \left(c'_{1,2}, c'_{2,3}, \ldots, c'_{n-1,n}\right) \tag{74}.$$

If all elements of this vector are equal, but the next element $c'_{n,n+1}$ has a different value, i.e., $c'_{n-1,n} \neq c'_{n,n+1}$, then the structural element $s_n$ is the final critical point of the substructure that is stable with respect to the parameter $u_k$, reduced to a qualitative value $u'_k$. In this case, we can perform the structural-parametric reduction $R_{sp}$:

$$S' = R_{sp}(S) = [s_1(u'_{1,k})]c'_{1,n}[s_n(u'_{n,k})] \qquad (75).$$

Indeed, if the relationship parameters $c'_{i,i+1}$ of the structural elements $s_i$ and $s_{i+1}$ are equal, then, according to the condition of interdependence between the energy of units and the energy of their relationships (63), the reduced parameter values of the structural elements (unit energy parameters) $u'_{i,k}$ and $u'_{i+1,k}$ will also be equal. Thus, we can simplify the structure $S$ to the structure $S'$ (75) in terms of the parameter $u'_k$ by eliminating repetitive structural elements, i.e., by compressing the structure, which corresponds to satisfying the conditions for reducing effective energy (65, 66).

However, for large differences (or distances) in the quantitative values of the parameters $u_{1,k}$ and $u_{n,k}$, significant information about the magnitude of their change relative to each other may be lost. To resolve this issue, it is necessary to introduce qualitative discrete thresholds for the quantitative change of this parameter according to a given scale or coordinate system.

For example, when considering a parameter such as the orientation vector of a segment or its position in a rectangular coordinate system, we can segment this coordinate system into multiple generalized (qualitative) substructures: half-planes, quadrants, or segments of quadrants. In this case, this segmentation will define the threshold values of orientation or position parameters, i.e., it will determine the limits of changes in quantitative parameters and the critical structural points as points of overcoming these threshold values.

Then, for equation (75), we need to define constraints for the parameters $u_{1,k}$ and $u_{n,k}$:

$$u_{1,k} \geq Tr_j, \quad u_{n,k} < Tr_{j+1} \tag{76},$$

where $Tr_j$ and $Tr_{j+1}$ are threshold parameter values that determine their belonging to the same segment of the scale or coordinate system. In this case, if $u_{n,k} \geq Tr_{j+1}$, a critical point is determined in the structure $S$ even with an unchanged value of the parameters $c'_{i,i+1}$.

Transitions between segments of scales or coordinate systems of different levels of generality, for example, increasing the generality of segments in a rectangular coordinate system from quadrant segments to quadrants and half-planes, also determine the steps of reducing the internal energy of the system—descending along the surface of its energy landscape.

The determination of threshold values, as well as the segmentation of scales or coordinate systems itself, should be carried out during the adaptation (training) process of the system for detecting (measuring) the parameters of perceived energy.

It is evident that by using the gradient of the second derivative, we transition to another qualitative scale, analyzing the acceleration or deceleration of parameter changes.

It is evident that there can be multiple qualitative (derivative) constant parameters that define stable substructures. In this case, a variety of different parametric relationships between the sequence of structural elements is formed. The establishment (selection) of the principal parameters for structural-parametric reduction is then carried out based on weight coefficients www, which are formed during the training process (43-51), i.e., during the structural reduction $R_w$.

Thus, the following sequence of internal energy reductions in the system must be performed: parametric reduction $R_{u,c}$, structural-parametric reduction $R_{sp}$, and structural reduction $R_w$:

$$E_{red} = R_w(R_{sp}(R_{u,c}(E_{total}^{(0)}))) < E_{total}^{(0)} \tag{77},$$

where $E_{total}^{(0)}$ is the total energy of the system at the initial structural level.

At the same time, transitions from quantitative to qualitative parameters, scales (coordinate systems), and their segments define the steps of the gradient descent along the surface of the system's energy landscape. The reinforcement of these scales is possible when considering the initial quantitative values of unit energy parameter differences (structural elements) not in an interval scale but in a ratio scale, where the parameters of the initial critical point determine an absolute value—the center of the coordinate system—and the parameters of the selected structural point are compared to this center.

Thus, the set of structural elements (units), their relationships, parameters, coordinate systems, scales, and segmentation thresholds, as well as critical structural points (bifurcation points), form the phase space of the given energy system. The reduced structure of units and their relationships reflects the evolution of the system's energy landscape during the learning process and represents an energy and structural attractor— a concept for a certain set (class) of structural measurements of the input energy parameters.

It follows that equation (77) represents a generalized form of the energy function of a self-organizing and evolving system in the form of a Lyapunov function.

## 6.3. Self-Organization and Evolution of the System Based on Energy

The considered reduction stages represent the process of forming a *stable energy structure—an attractor*—that corresponds to the most general and invariant structure of a certain class of input information.

*Thus, the obtained reduced structure—the attractor—serves as an internal model of a certain class of the spatiotemporal structure of the measured input energy parameters. It defines the set of available microstates of the system, which correspond to a given*

*macrostate determined by the system's response. We will call this internal model the concept of a given class.*

A similar scheme is implemented in quantum systems using a family of Hamiltonians:

$$H(\lambda) = H_{units} + \lambda V \tag{78}.$$

Here, $H_{units}$ represents the part of the Hamiltonian describing the internal (or "unit") properties of the system without considering interactions, while $\lambda V$ represents the interaction, whose strength is regulated by the parameter $\lambda$. According to the Hellmann-Feynman theorem [34], if $\Psi(\lambda)$ is a normalized eigenstate of $H(\lambda)$ with energy $E(\lambda)$, then:

$$\frac{\partial E}{\partial \lambda} = \langle \Psi(\lambda) \left| \frac{\partial H}{\partial \lambda} \right| \Psi(\lambda) \rangle \tag{79}.$$

Since the dependence of the Hamiltonian on $\lambda$ is given by $\lambda V$, we have $\frac{\partial H}{\partial \lambda} = V$, and accordingly,

$$\frac{\partial E}{\partial \lambda} = \langle \Psi(\lambda) | V | \Psi(\lambda) \rangle \tag{80}.$$

If the operator $V$ describes an attractive (energy-compressing) interaction, then its expectation value $\langle \Psi(\lambda)|V|\Psi(\lambda) \rangle$ has a negative value. This means that as $\lambda$ increases (i.e., as the interaction strengthens), the system's energy $E(\lambda)$ decreases (becomes more negative).

It is important to note that energy reduction does not occur solely due to contributions from $\lambda V$. The wave function $\Psi(\lambda)$ also changes to optimize (minimize) the total energy. As a result, the adaptation of $\Psi(\lambda)$ leads to a decrease in the contribution from $H_{units}$ as well. In other words, the system adjusts its internal state in such a way that, along with the strengthening of the attractive interaction, it reaches a lower energy state.

Consequently, quantum systems exhibit the same interdependence: the establishment of favorable (energy-reducing) interactions leads to a decrease in both the energy of their interactions and the energy of individual units.

In statistical physics, the probability that a system is in a state with energy $E$ is determined by the Boltzmann distribution (20). This implies that at a given temperature, states with lower energy are significantly more probable. In thermodynamic equilibrium, the system almost always settles into a configuration corresponding to the minimum free energy, ensuring maximum statistical probability and stability.

Thus, in the considered framework, the energy reduction equation of the system (77), along with the equations for effective energy (63, 64) and the system's total energy function $E_{total}$ (51), not only illustrate how the system's energy decreases but also fundamentally define the dynamics of its minimization process. All these equations function as a Lyapunov function: they monotonically decrease during the system's evolution, directing it toward a stable, energy-optimal state.

Unlike the energy function in the Hopfield model, which is used merely to demonstrate the convergence of dynamics, these energy functions directly govern the reduction process due to the following features:

- Interdependence of parameters: Since the internal energies of the units and the energy of their interconnections are mutually dependent (conditions (53, 60–62)), changes in the interconnection parameters $C$ compel the system to adjust the parameters $U$ to minimize $E_{units}$. This ensures a coordinated decrease in the total energy $E_{total}$ (51).

- Direct control of dynamics through parametric reduction: Due to equilibrium conditions (60-62), minimizing $E_{eff}(C)$ (64) is not merely a demonstration but a mechanism by which the system "self-regulates." Changes in interactions lead to the optimization of both the structural unit energies and their interconnections.

- Direct control of dynamics through structural-parametric reduction: When conditions (71-73, 76) are met, the system evolves toward stable substructures (75) with minimal energy.

- Direct control of dynamics through structural reduction: When conditions (46-50) are met, the system reduces its total energy (51) by optimizing its structure (52).

Thus, equation (77) describes the general concept of a self-organizing and evolving energy system, where the system's order, based on minimizing its internal energy, determines its evolution toward a stable state—an attractor—analogous to processes observed in physical or quantum systems.

*In this concept, the stages (steps) of reduction limit the number of microstates available to the system for a given macrostate*. This allows the system to "abstract" from microstates (energy elements and connections) that exert an insignificant or unstable influence (e.g., noise) on achieving the target macrostate. From a thermodynamic perspective, this means a reduction in the system's accessible phase space.

However, this is possible only for a singular (non-generalized) energy landscape, i.e., one with a single macrostate. Consequently, evolutionary systems (such as evolutionary artificial neural systems) must construct generalized energy landscapes through interconnections between individual energy landscapes, as occurs in complex physical systems, rather than by simply merging and averaging them, as is done in modern connectionist ANNs. This concept of interconnections between individual neurons (e.g., based on competitive mechanisms), which respond to holistic patterns, is well known in neuroscience as the theory of single neurons or "grandmother cells" [35-37].

Now, let us consider the thermodynamics of the structural reduction process, where the key role is played by the values of the weight coefficients www. If $S$ is the Boltzmann entropy and $\Omega$ is the number of accessible microstates, then the weight coefficients $w$ constrain their quantity

$$\Omega(w) = \Omega_0 w^\gamma, \gamma > 0 \qquad (81),$$

where $\Omega(w)$ is the number of accessible microstates for a given macrostate of the system at parameter $w$;

$\Omega_0$ is the initial (total) number of microstates that the system could have at $w = 1$;

$w^\gamma$ represents the nonlinear dependence of the number of microstates on $w$, where $\gamma > 0$ is a positive parameter regulating the rate of decrease in the number of microstates as $w$ increases. In real physical systems, the number of accessible microstates typically does not change linearly with structural modifications.

Then, the entropy of the system takes the following form:

$$S(w) = k_B \ln \Omega_0 + k_B \gamma \ln w \qquad (82).$$

Thus, an increase in $w$ leads to a local reduction in thermodynamic entropy for the given macrostates of the system, which also serves as confirmation of our previous conclusions.

Now, we can express $E_{total}$ as:

$$E_{total} = E_0 - TS(w) \qquad (83),$$

where $E_0$ is the initial energy of the system (prior to training) or its energy at maximum entropy ($w = 1$), when all possible microstates are accessible;

$T$ represents the temperature (or an analogous parameter characterizing the degree of random fluctuations).

Then,

$$E_{total} = E_0 - k_B \ln \Omega_0 - k_B \gamma \ln w \qquad (84),$$

and

$$\frac{dE_{total}}{dw} = -\frac{k_B T \gamma}{w} < 0, \quad when \ w > 0 \qquad (85).$$

That is, energy indeed decreases as www increases due to the reduction of the phase space of accessible microstates.

However, restricting the number of microstates at the stage of structural reduction (52) does not simply mean "removing" elements or connections from the system's structure. In physical systems, such constraints must lead to a restructuring of the system's internal organization. This restructuring is crucial because the system must remain coherent and maintain its macroscopic order. That is, the system must *evolve* through the reorganization of its structural elements $E_i$ and connections $E_{ij}$ so that the new microstates available after the reduction still support the macroscopic state of the system. As the number of accessible microstates decreases, the system must optimize the connections between elements to reduce entropy and preserve the integrity of the concept's structure.

Thus, for the self-organization and evolution of an energy-based system, the following principles must be satisfied:

- *Energy minimization rules*, where the system must tend toward a state in which its internal energy, both of the elements and their interactions, is minimized under the constraints defined by the parameters www and the gradient descent along their energy landscapes;

- *Structural optimization rules*, where the system must not only minimize its energy but also optimize its structure, not only retaining the connections and elements crucial for maintaining the macrostate but also forming new, more stable connections between elements to establish the concept.

At the same time, several principles must be followed:

- *The principle of local energy minimization*, where each element of the system must strive to minimize its own energy while considering its surrounding connections. This should occur as a result of modifying interactions with neighboring elements, leading to a transition into a more stable state. This principle can be expressed as:

$$\frac{dE_i}{dx_i} = 0 \tag{86},$$

where $x_i$ are parameters describing the position or state of an element.

- *The principle of entropy minimization*, where, in order for the system to remain structured, it must maintain a maximally ordered state at minimal entropy. This principle reflects the system's tendency to reduce unstable states (e.g., irrelevant connections or elements) that may cause fluctuations or disturbances. In the case of the considered concept with weight coefficients www, which reduce the number of accessible microstates, the entropy minimization principle can be expressed as:

$$\frac{dS}{dw} < 0 \tag{87};$$

- *The principle of maximizing connectivity (or topological stability)*, where, in order for the system's structure to remain intact and stable, elements must preserve the overall network topology even if certain connections or parameters are removed. This means that the system should not "fall apart" or lose its key functions or qualities. Mathematically, this can be related to the concept of topological stability, where the system must avoid situations in which its structure becomes fragmented. This principle can be implemented using graphs or networks.

Thus, we observe a fundamentally different role of weight coefficients in the energy-based model of an ANN compared to the classical model, where weight coefficients define the depth of potential "wells" of units or their interconnections in a resting state. In the considered model, the formation of internal energy of units and their interconnections must be based on external energy rather than on "pseudo-energy" (e.g., error energy), as is the case in modern connectionist ANNs. At the same time, the construction of the structure of this internal energy and its minimization is carried out based on reduction rules governed by energy functions that model the conditions of energetic interactions (interdependencies). This model is inspired by energetic processes occurring in physical systems.

*General conclusions regarding the processes of self-organization and system evolution*:

1. Self-organization (as the formation of an energy landscape) and evolution (as the change of an energy landscape) of any energy-based system are possible only through the transformation of external energy into the system's internal energy and the establishment of energy interconnections between the internal energies of individual units. If the units and their interconnections do not possess a coherent internal energy structure, then it is evident that such self-organization based on the minimization of the system's internal energy is impossible.

2. For any energy-based system that has a unified structure of the internal energy of units and their interconnections, there exists an energy function that determines energy minimization.

The basis of this assertion is the fact that if we consider systems for which a smooth functional describing energy can be constructed, and if the dynamics of this system have a gradient nature or can be reduced to a process accompanied by a monotonic decrease in energy, then for such systems, a Lyapunov function can always be constructed and interpreted as a function of energy.

We may not know this function if the energy structure is unknown to us, as is the case in Prigogine's dissipative systems. In that case, we can apply another mechanism to analyze energy minimization processes. However, if we are designing a deterministic system (e.g., an artificial neural network) that is not solely based on a probabilistic or statistical approach, then to organize self-structuring processes, we require a unified internal energy structure.

3. Regarding alternative approaches to "self-organization" processes, such as weight coefficient adaptation in ANNs, this "self-structuring" process is conditional. Its implementation requires the participation of an *external system* (a "Maxwell's Demon") in one form or another. In this case, it is precisely the optimized parameters of external "pseudo-energy" (e.g., error energy) that shape the internal energy structure of units (neurons), as seen in classifying ANNs, or the energy of interconnections between units, as seen in generative ANNs.

4. The principle of internal energy minimization is the foundation of self-organization and evolution not only of any physical system but also of any informational system, provided that "internal energy" is understood as any quantity characterizing the system's state and subject to optimization. Ultimately, self-organization is an optimization process, and the chosen interpretation of "energy" determines how this process is described in a given context.

5. A key constraint in ANN self-organization processes is the presence of a "Maxwell's Demon," which is necessary for minimizing external "pseudo-energy." The internal energy structure of the system, shaped under the influence of "pseudo-energy" parameters, does not fully reflect the measurable parameters of the input energy, i.e., the information at the system's input. This leads to a situation where "pseudo-energy" parameters influence either the formation of the internal energy of units in first-type ANNs or the formation of the internal energy of interconnections between units in second-type ANNs.

In contrast, when a system self-organizes based on internal energy formed from input information, we can eliminate the need for a "Demon" and evolve the energy landscape in a resting state. In such a process, it does not matter how external energy is physically represented; what matters is the information the system can extract from this energy to form its internal energy.

## 6.4. Internal Model of the External World. Information

It can be stated that the internal structure of a self-organizing energy-based system represents a *model of the system's external world*, denoted as *M*. This model is formed through the following processes:

- Perception of structured external energy (detection and measurement of its parameters) and its transformation into internal energy.

- Formation (self-organization) of the system's initial energy landscape structure in a resting state.

- Reduction (evolution) of this structure during the learning process.

Mathematically, we define the internal model as:

$$M = R(S_{int}(P(E_{int})) = Con \qquad (88),$$

$$P(E_{int}) = D(S_{ext}(E_{ext})) \qquad (89),$$

where:

*R* is a generalized reduction operator;

$S_{int}(P(E_{int}))$ represents the structure of the set of parameters *P* of the system's internal energy $E_{int}$;

*Con* is an attractor of the system for a given class of input information vectors;

*D* is a generalized detection (measurement) operator;

$S_{ext}$ represents the structured (spatiotemporal) parameters of external energy $E_{ext}$.

The model *M* can be represented as a hypergraph, while its evolution can be described as a hypernetwork [38]. This representation enables the formal proof of the uniqueness of the concept *Con* (the attractor) for a given class of input information vectors. Additionally, it ensures parametric and structural stability as well as the convergence of the initial structure to the attractor within a finite number of reduction steps. It is likely that biological neurons, such as pyramidal neurons, implement the energy reduction process in their dendritic structures in a manner similar to how it may occur through the use of hypergraphs [39].

If we analyze the process of interpreting input energy using this model and the formation of output energy (or information), we can draw an analogy with the inference process in ANNs. During interpretation, the system perceives input energy and constructs its current model $M'(t)$, which is then reduced and compared with the stored model $M$ (the attractor or *Con*):

$$M'(t) = S_{int}(P(E_{int}(t)) \tag{90}.$$

Now, let us represent these models as hypergraphs: $G' = R(M'(t))$ and $G = M$.

If $G'$ is isomorphic to $G$, we define a bijection function:

$$f: G' \rightarrow G \tag{91},$$

where $f$ is a one-to-one mapping. In this case, the reduction process converges to the stored model: $R(M'(t)) = M$ and $M' \in M$.

If $G$ is a subgraph of $G'$ (subgraph isomorphism), then there exists an injective function:

$$f': G \rightarrow G' \tag{92}.$$

In the case of

$$f': G' \rightarrow G \tag{93},$$

we are dealing with associative recognition.

In this case, the interpretation condition $M' \in M$ still holds, and we can describe the process of interpretation as:

$$M = Int(M') \qquad (94),$$

where *Int* is the generalized interpretation operator that maps the model $M'$ to the stored model $M$.

However, if conditions (91-93) are not satisfied, then $M'$ cannot be interpreted by the stored model $M$. This means that the structure and parameter values of the input energy (input information) do not match the stored energy (the system's energy landscape in its resting state).

As a result of the interpretation of the model $M'$, a response is formed—either the system's output information $I$ or its output energy, represented as the structure of parameters of the emitted energy $P(E_{out})$, which correspond to the respective parameters of the system's internal energy. The formation of this parameter structure at the system's output is governed by the activation function (or reading function) $F$:

$$F(M) = S_{out}(P(E_{out})) = I \qquad (95),$$

where $S_{out}$ represents the structure (e.g., spatial-temporal) of the parameters of external energy $P(E_{out})$. These conclusions align with the interpretation (33) of Gibbs free energy as the structured portion of the system's output energy (enthalpy). This is further confirmed by the consistent behavior of informational and thermodynamic entropy, both at the level of an individual neuron model and within an ANN as a whole.

The system's response $I$ may be:

- Constant (unchanging) for a specific class of input energy under $F(M)$, meaning it depends on the structure of the concept;

- Dynamic under $F(M')$, meaning the system's response depends on the current state of its energy landscape.

For example, in the simplest case, a constant system response may reflect the sum of the structural elements of the concept. Such a response enables the realization of a "Winner-Take-All" (WTA) competition process among the responses of neighboring concepts, represented by either artificial or biological neurons. Different types of system responses, associated with satisfying conditions (91–93), may define various types of biological neuron activity and the corresponding interpretations of their "neural code" [40].

Thus, based on the conducted research, a generalized formal definition of information can be formulated.

*Definition:* *The structure of* interdependent *subjectively perceived and measurable parameters of external energy, which serves to form the structure of interconnected parameters of the system's internal energy—its model of the external world, or to interpret it using an already existing model, constitutes information for this system.*

If a system perceives external energy as conditional energy in the form of a structure composed of elements of its internal alphabet (e.g., a vector of input signals), as observed in modern information systems and ANNs, then its model of the external world, as well as the process of signal vector interpretation, can only be constructed based on the statistical relationships between the elements of this alphabet. This leads to the understanding of information in terms of Shannon entropy.

This definition of information allows us to consider *informational diversity of models* $Div(I)$ as the difference between models M′ and M under the conditions (91) or (92):

$$Div(I) = M' - M \tag{96},$$

A possible metric for this diversity could be, for instance, the edit distance between graphs.

An energy system described by equations (88), (89), and (95) can be classified as a *system with an evolutionary energy landscape structure that self-organizes based on information*.

In such a system, internal energy is minimized due to the "compression" of the energy landscape structure. This suggests that the system's response (95) can be interpreted as encoding information about the state of the external world model. In complex information-energy systems, such as the brain, the necessity of such encoding arises from the need for further information processing by interconnected models of the external and internal world. Within this interpretation, it is more accurate to consider the examined system not as a "complete" model of the external world, but rather as a functional fragment of it.

Furthermore, such "encoded" information at the output of the examined system will be perceived only by a system equipped with specific sensors (detectors), similar to how information exchange occurs between biological neurons. This reduces the universality of information exchange between autonomous systems.

To enable a universal process of information exchange between autonomous systems with similar models of the external world—for instance, between humans—there is a need for an information "decoding" system, i.e., a mechanism of divergence, which we can conditionally compare to a generative language model of ANNs.

For example, suppose a person sees a cat. In the sequence of successive activations of neuron groups in their brain, the "Cat" class holistic neuron is activated. However, to convey this information to another person, it is necessary to decode this reaction into an informational sequence, such as a sequence of sounds (spoken word) or hand movements (writing the word "Cat").

It is important to note that the variety of internal energy types in a system depends on the number and types of sensors (detectors) of external energy. Thus, it is sensory systems that create the multimodality of internal energy and define the diversity of the phase space of the energy system.

For artificial information-energy systems, the interpretation of external and internal energy depends on the tasks these systems are designed to solve. This perspective differs from the modern multimodality concept in ANNs, where different types of information (visual, auditory, textual) are considered modalities. In such ANNs, it would be more accurate to refer to this property as multi-agent interaction rather than multimodality.